\documentclass[lettersize,journal]{IEEEtran}

\usepackage{amsmath,amsfonts}
\usepackage{algorithmic}
\usepackage{algorithm}
\usepackage{array}
\usepackage[caption=false,font=normalsize,labelfont=sf,textfont=sf]{subfig}
\usepackage{textcomp}
\usepackage{stfloats}
\usepackage{url}
\usepackage{xcolor}
\usepackage{color}
\usepackage{verbatim}
\usepackage{graphicx}
\usepackage{cite}
\usepackage{multicol}
\usepackage{multirow}
\usepackage{booktabs}
\usepackage{caption}
\usepackage{amsmath}
\usepackage{bm}
\usepackage{bbding}
\usepackage{cancel}

\begin{document}

%\title{Gaussian Splatting SLAM using a Rotating Device and Multiple RGB-D Cameras}
\title{Robust Gaussian Splatting SLAM by Leveraging Loop Closure}
\author{Zunjie~Zhu, Youxu Fang, Xin Li, Chengang Yan, Feng Xu, Chau Yuen,~\IEEEmembership{Fellow,~IEEE},  Yanyan Li
        % <-this % stops a space
\thanks{Z. Zhu and Y. Fang are with the School of Communication Engineering, Hangzhou Dianzi University, Hangzhou, China. Z. Zhu and Y. Fang are also with the Lishui Institute of Hangzhou Dianzi University. E-mail: \{zunjiezhu, fangyouxu\}@hdu.edu.cn.}
\thanks{C. Yan is with the School of Automation, Hangzhou Dianzi University, Hangzhou, China. C. Yan is also with the Faculty of Applied Sciences, Macao Polytechnic University, Macao, China. E-mail: cgyan@hdu.edu.cn.}
\thanks{X. Li and C. Yuen are with School of Electrical and Electronics Engineering, Nanyang Technological University, Singapore. E-mail: \{li-x, chau.yuen\}@ntu.edu.sg.}
\thanks{F. Xu is with the BNRist and School of Software, Tsinghua University, Beijing, China. E-mail: feng-xu@tsinghua.edu.cn.}
\thanks{Y. Li is with the Technical University of Munich, Munich, Germany. Y. Li is the corresponding author. E-mail: yanyan.li@tum.de.}
}


\markboth{Journal of \LaTeX\ Class Files,~Vol.~14, No.~8, August~2021}%
{Shell \MakeLowercase{\textit{et al.}}: Robust Gaussian Splatting SLAM by Leveraging Loop Closure}

%\IEEEpubid{0000--0000/00\Additionally, a local bundle adjustment scheme is implemented to further optimize camera poses using photometric and geometric constraints.}


\maketitle

\begin{abstract}
% background and problem statement
3D Gaussian Splatting algorithms excel in novel view rendering applications and have been adapted to extend
the capabilities of traditional SLAM systems. However, current Gaussian Splatting SLAM methods, designed mainly for hand-held RGB or RGB-D sensors, struggle with tracking drifts when used with rotating RGB-D camera setups. 
% Solution of this paper
In this paper, we propose a robust Gaussian Splatting SLAM architecture that utilizes inputs from rotating multiple RGB-D cameras to achieve accurate localization and photorealistic rendering performance. The carefully designed Gaussian Splatting Loop Closure module effectively addresses the issue of accumulated tracking and mapping errors found in conventional Gaussian Splatting SLAM systems. 
% Contribution 1
First, each Gaussian is associated with an anchor frame and categorized as historical or novel based on its timestamp. By rendering different types of Gaussians at the same viewpoint, the proposed loop detection strategy considers both co-visibility relationships and distinct rendering outcomes.
% Contribution 2
Furthermore, a loop closure optimization approach is proposed to remove camera pose drift and maintain the high quality of 3D Gaussian models. The approach uses a lightweight pose graph optimization algorithm to correct pose drift and updates Gaussians based on the optimized poses. Additionally, a bundle adjustment scheme further refines camera poses using photometric and geometric constraints, ultimately enhancing the global consistency of scenarios.
% Contribution 3
Quantitative and qualitative evaluations on both synthetic and real-world datasets demonstrate that our method outperforms state-of-the-art methods in camera pose estimation and novel view rendering tasks. The code will be open-sourced for the community.
\end{abstract}


\begin{figure}[htbp]
   \centering
    \captionsetup[subfloat]{labelformat=empty}
	\subfloat[w/o Loop]{
    \begin{minipage}[b]{0.5\linewidth}
    \includegraphics[width=1\linewidth]{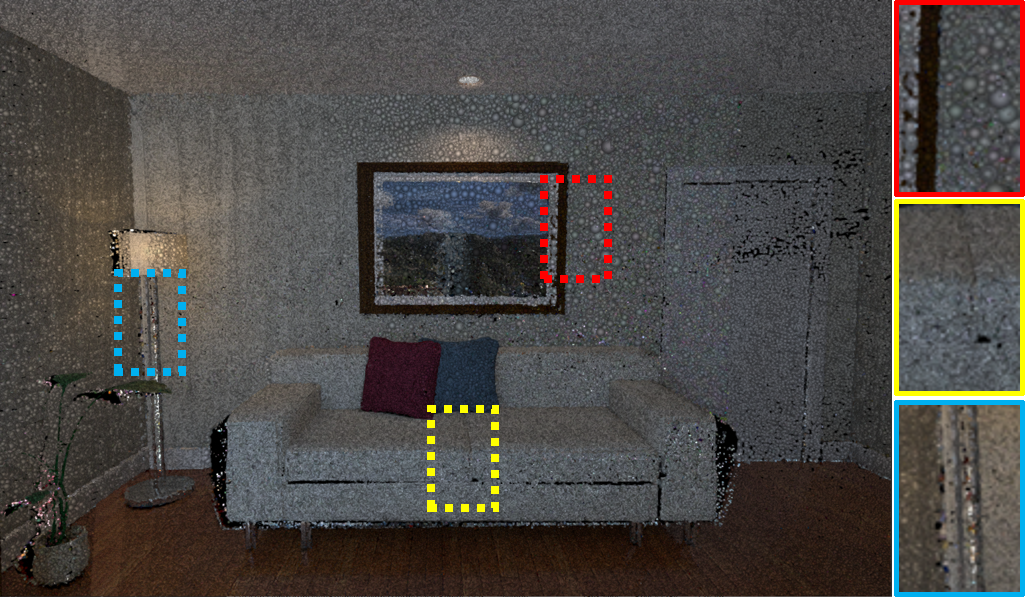}\vspace{4pt}
    \includegraphics[width=1\linewidth]{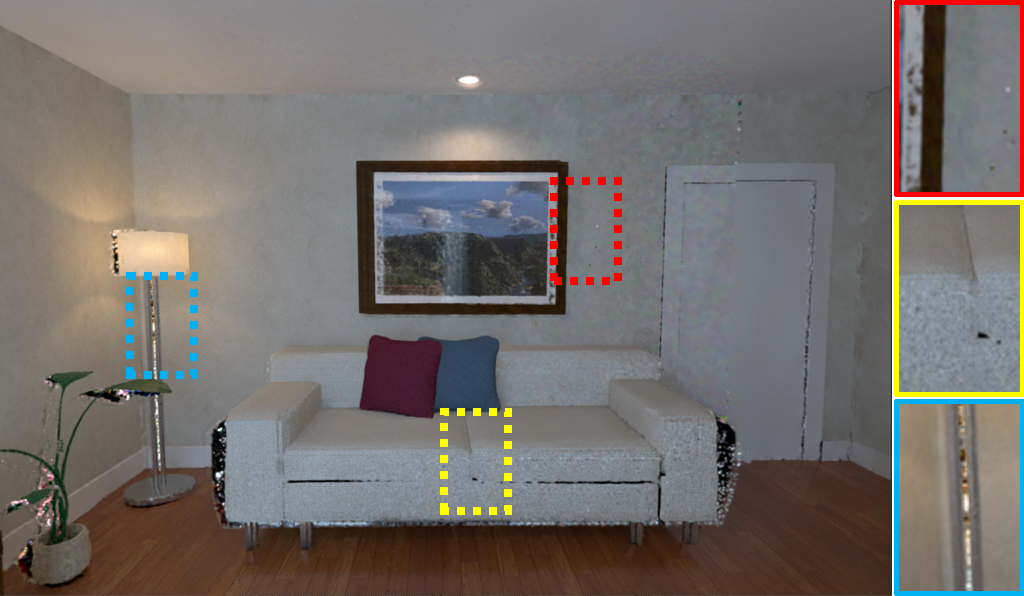}
    \end{minipage}}
	\subfloat[Loop]{
    \begin{minipage}[b]{0.5\linewidth}
    \includegraphics[width=1\linewidth]{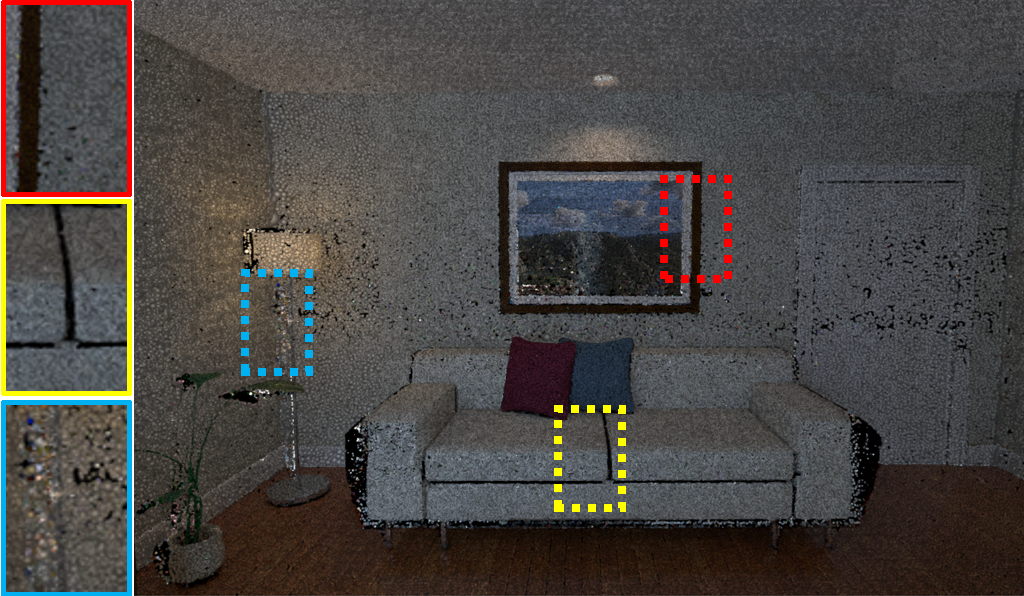}\vspace{4pt}
    \includegraphics[width=1\linewidth]{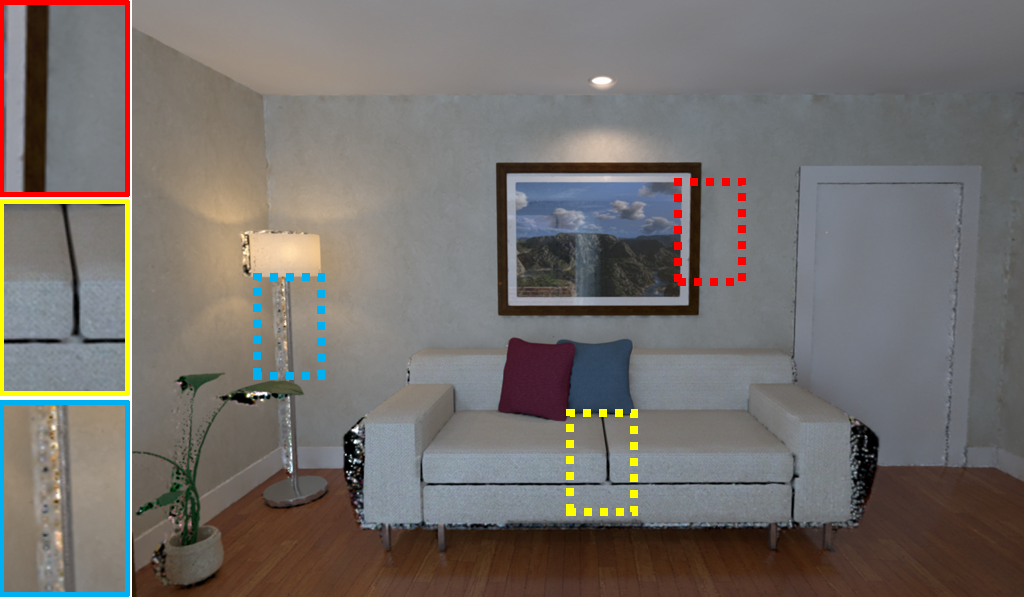}
    \end{minipage}}
	\caption{\textbf{Example of loop closure optimization:} A comparison of our method without loop optimization (left) and with loop optimization (right) on 3D Gaussian ellipsoid visualization (top) and novel view rendering (bottom).}
	\label{fig}
\end{figure}

\section{Introduction}
\IEEEPARstart{S}{imultaneous} localization and mapping (SLAM) is a foundational technology in robotics and computer vision, enabling smart robots to navigate and understand their environments by reconstructing a map and simultaneously localizing themselves within it. Although general SLAM systems utilizing monocular~\cite{mur2015orb,li2020structure}, stereo~\cite{sun2018robust,engel2015large}, RGB-D~\cite{mur2017orb,newcombe2011kinectfusion}, and visual-inertial~\cite{qin2018vins,campos2021orb} sensors have shown impressive performance in camera pose tracking and 3D reconstruction, they face challenges in rendering novel photorealistic images that are crucial for advancing further intelligence applications.
% % Background
% 3D Gaussian Splatting algorithms have demonstrated impressive performance in novel view rendering tasks, which have been adapted for the SLAM domain to enhance the representation capabilities of traditional SLAM systems.
% % Problems
% However, these Gaussian Splatting SLAM systems are primarily designed for handheld RGB or RGB-D sensors and suffer from degeneration issues, particularly with rotating sensor devices equipped with multiple RGB-D cameras, leading to tracking drifts.

To enhance the novel view rendering capabilities of SLAM systems, neural radiance fields~\cite{mildenhall2021nerf} using a single Multi-Layer Perceptron (MLP) as a scene representation have been integrated into conventional tracking and mapping methods to build NeRF-SLAM~\cite{rosinol2023nerf,chung2023orbeez}. In pursuit of rendering efficiency and quality, point-based explicit representations, such as 3D Gaussian Splatting (GS) algorithms, including 3D-GS~\cite{kerbl20233d}, 2D-GS~\cite{huang20242d}, and Gaussian surfel~\cite{dai2024high}, are employed in novel view synthesis tasks. Through the process of GS optimization, camera poses and 3D Gaussians are incrementally estimated in GS-based SLAM systems~\cite{matsuki2024gaussian,yan2024gs}. Unlike traditional SLAM approaches that typically rely on features for camera pose tracking, GS-based SLAM methods fix the parameters of 3D Gaussians and optimize the camera poses of the current frame by rendering Gaussians to images to compute photometric~\cite{kerbl20233d} and geometric~\cite{li2024geogaussian} residuals between observed and rendered images. However, these GS-based SLAM systems still face challenges, such as camera drift during the tracking process, despite the implementation of bundle adjustment modules to refine the 3D Gaussians and keyframe camera poses. Moreover, these algorithms are primarily designed for traditional handheld sensor setups, which limits their applicability in more advanced SLAM systems, particularly those involving rotating devices~\cite{10582466} with multiple RGB-D cameras.

In this paper, we introduce a robust Gaussian Splatting SLAM architecture designed to address typical degradation problems in pose estimation and scene synthesis using rotated multiple RGB-D cameras. By integrating a novel GS-based loop closure module which includes both loop detection and drift removal strategies, our system achieves accurate camera localization and photo-realistic view rendering performance. 
First, each Gaussian is associated with an anchor frame based on its generation timestamp. By categorizing 3D Gaussians into history and novel groups based on timestamps, we propose a novel loop detection strategy that leverages both the number of co-visible observations and the rendering distances of different Gaussians.
Furthermore, we introduce a loop closure optimization approach to address camera pose drift and preserve the high quality of 3D Gaussian models. This approach employs a lightweight pose graph optimization module to correct pose drift and updates Gaussians based on the refined poses of their anchor frames. Additionally, a bundle adjustment scheme further optimizes camera poses using photometric and geometric constraints to achieve global consistency between viewpoints and 3D Gaussians.
The contributions of this work are summarized as follows:
\begin{enumerate}
\item We categorize 3D Gaussians into novel and history groups based on their generation timestamps, leveraging this classification for enhanced loop detection.
\item  We propose a loop closure optimization to correct camera pose drift and dynamic updates to Gaussian submaps, preserving model integrity.
\item We propose a two-stage bundled adjustment strategy that uses constraints on the rendered image and pose graph structure to globally refine the camera pose.
\end{enumerate}
Through extensive quantitative and qualitative evaluations on both virtual and real-world datasets, we demonstrate that our Gaussian Splatting SLAM system significantly outperforms state-of-the-art methods in both camera pose estimation and novel view rendering tasks. The open sourcing of our code will further enable the community to build upon and extend this work.

\section{Related Work}
\subsection{Visual-based Tracking and Reconstruction}
Given monocular~\cite{mur2015orb,engel2014lsd}, stereo~\cite{mur2017orb,engel2015large}, or RGB-D~\cite{li2021rgb,schops2019bad} videos, the goals of visual SLAM methods are to estimate 6-DoF camera poses and reconstruct unknown scenes incrementally. 
Feature-based SLAM methods~\cite{mur2015orb,mur2017orb,campos2021orb} rely heavily on extracting sparse points features from images, which are fed to the feature matching algorithms to estimate the relative motions between viewpoints. To reduce the computation in descriptor calculation, FastORB-SLAM~\cite{fu2021fast} proposes a lightweight and effective method to track key points without computing descriptors.
In man-made scenes, low-textured scenarios are difficult to provide enough reliable points to achieve robust tracking and mapping. Therefore, Other types of features, such as lines~\cite{gomez2019pl} and planes~\cite{li2021rgb,gong2021planefusion}, are explored to compensate for the degeneration cases. Benefitting the matching results of planes and lines, new residuals are introduced in SLAM systems~\cite{rosinol2020kimera} to achieve more accurate and robust tracking results.
Instead of reconstructing sparse point clouds as these feature-based SLAM systems did, the advent of depth cameras supports efficient dense modeling. Among these, Kinectfusion~\cite{newcombe2011kinectfusion} enables real-time 3D model construction using only depth images, thus achieving real-time rigid reconstruction with consumer cameras. ElasticFusion~\cite{whelan2015elasticfusion} employs dense frame-to-model tracking, window-based surface fusion, and frequent model refinement through non-rigid surface deformation to achieve dense SLAM modeling. BundleFusion~\cite{dai2017bundlefusion} offers one of the best-performing algorithm frameworks, particularly excelling in loop closure and robust error correction in textured environments. Volume Fusion algorithms enable reconstruction in larger scenes and support real-time tracking and reconstruction using RGB-D cameras with CUDA acceleration. InfiniTAM~\cite{prisacariu2017infinitam} provides a lightweight yet powerful solution for 3D scene reconstruction and tracking on limited hardware resources. DynamicFusion~\cite{newcombe2015dynamicfusion} focuses on dense SLAM reconstruction of non-rigidly deforming scenes by calculating scene flow and inverse transformation. BAD-SLAM~\cite{schops2019bad} introduces a fast and direct bundle adjustment method capable of dense map optimization previously unachievable by traditional BA approaches. Even though these SLAM systems provide robust camera pose estimation and 3D scene reconstruction solutions, it is difficult to generate high-quality RGB and depth maps at novel viewpoints based on these 3D reconstructed models.

\subsection{Gaussian Splatting SLAM}

3D GS approaches are proposed for photo-realistic novel view rendering tasks based on ground truth camera poses and sparse point clouds provided by Structure-of-Motion algorithms~\cite{schonberger2016structure,cui2023mcsfm}. Compared to Neural Radiance Fields (NeRF)~\cite{mildenhall2021nerf} that uses implicit scene representation based on MLP networks, 3D GS~\cite{kerbl3Dgaussians} employs explicit representations of scenes and differentiable rendering modules. This ensures real-time rendering capabilities while introducing unprecedented levels of control and scene editing. The core concept of 3D GS involves constructing a large collection of 3D Gaussians that accurately capture the essence of a scene. This facilitates free-viewpoint rendering and is optimized through differentiable rendering to adapt to the textures present in a given scene.

In GS-SLAM systems, camera pose estimation, 3D mapping, and view rendering are considered in a unified architecture. Anisotropic Gaussians enable more detailed and accurate environmental modeling, particularly suitable for complex or dynamically changing scenes. The Gaussian Splatting SLAM (MonoGS)~\cite{matsuki2024gaussian} marks the first application of 3D GS for incremental tracking and rendering.
In response to challenges such as slow rendering speeds and difficulty in optimization when dealing with large-scale, high-density environments, a novel SLAM system, SplaTAM~\cite{keetha2024splatam} is introduced to utilize a silhouette mask to obtain the scene density, achieving faster rendering and optimization speeds while simplifying the dense mapping process. 
%MonoGS extends the applicability of 3DGS by utilizing monocular cameras for scene reconstruction. 
RGBD GS-ICP SLAM~\cite{ha2024rgbdgsicpslam} employs Generalized-ICP to estimate poses through matching Gaussians from the current frame and the map. Keyframe selection strategies further enhance rendering performance and tracking capabilities. These advancements highlight the versatility and effectiveness of 3D Gaussian Splatting across various SLAM applications, demonstrating its potential to handle complex environments and improve real-time rendering and tracking in dynamic scenes. However, these methods heavily depend on the (appearance and depth) alignment between observed and rendered images. When the residuals provide limited constraints for rendering-based optimization, both pose estimation and rendering performance degrade significantly. In contrast, the proposed method introduces a Gaussian Splatting Loop Closure module that effectively addresses errors in camera pose estimation and 3D Gaussian representation for rotating multiple RGB-D cameras, resulting in geometrically consistent 3D Gaussians.

%However, these methods heavily rely on the appearance and depth alignment between observed and rendered images, when the these residuals provide limited constraints for rendering-based optimization modules, the pose estimation and rendering performances degenerate significantly. Compared to these systems, the proposed method provides a Gaussian Splatting Loop Closure to remove errors in camera pose estimation and 3D Gaussian, obtaining geometrically inconsistent reconstructions of the scene.

\section{System Overview}
\begin{figure*}
    \centering
    \includegraphics[width=\linewidth,]{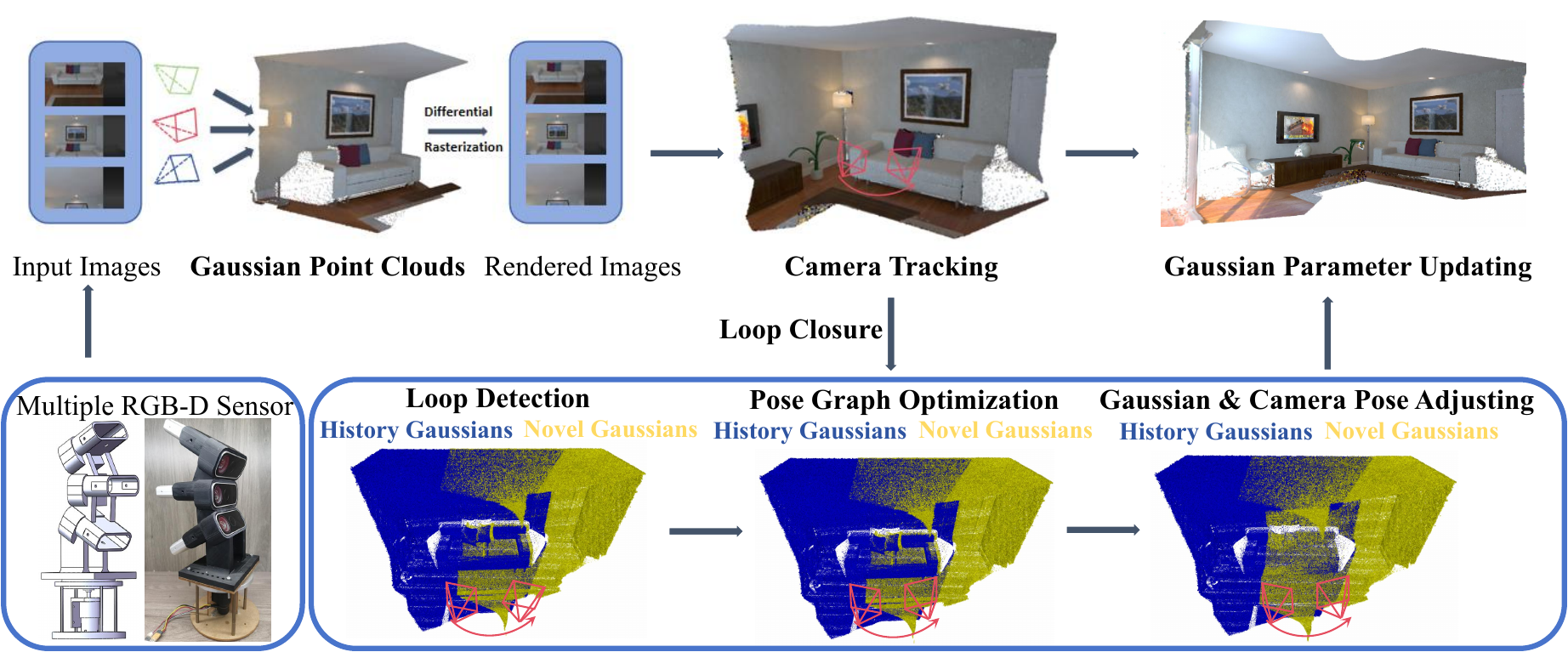}
    \caption{\textbf{Architecture of the proposed Gaussian Splatting SLAM.} The input of our system is the current RGB-D frame from rotating multiple RGB-D cameras. In the camera tracking and Gaussian parameters update process, we utilize differential rasterization results of three cameras to design effective loss functions. If a loop is detected, pose graph optimization is triggered first, then 3D Gaussian positions will be adjusted based on updates of camera poses, and finally a local bundle adjustment module is employed to further refine camera poses, ultimately achieving accurate camera poses and a 3D Gaussian map. }
    \label{fig:architecture}
\end{figure*}

\subsection{Rotating Multiple RGB-D Cameras}
%Compared to handheld scanning, fixed scanning~\cite{yang2020noise} scans an indoor scene by rotating the camera in a fixed place using a robot as shown in Figure~\ref{fig:architecture}. Additionally, there are overlaps between these three cameras, and the whole view field is extended largely to scan the unknown environments. Therefore, there are several advantages of using rotating device with RGB-D cameras in unknown scene reconstruction and mapping tasks. First, the camera motion of the type of device is highly controllable, more computation energies can be used in mapping and rendering modules. Furthermore, the large view fields make scanning more efficient. And the optimization module can be implemented since loop is generated in a short time.

Compared to handheld scanning, fixed scanning~\cite{yang2020noise} involves using a robot to rotate the camera setup in a fixed position, as illustrated in Figure~\ref{fig:architecture}. This setup~\cite{10582466} typically includes multiple RGB-D cameras with overlapping fields of view, significantly extending the overall coverage and enabling a comprehensive scan of unknown environments.  There are several advantages to using a rotating device with RGB-D cameras for scene reconstruction and rendering tasks in unknown environments. 
First, the robot precisely controls the camera motion, eliminating inconsistencies and errors associated with handheld scanning. This controlled movement enables more accurate data collection and alignment, resulting in higher-quality reconstruction performance. Additionally, by handling camera motion, computational resources can be allocated more efficiently to the mapping and rendering modules, enhancing their performance for faster and more accurate scene rendering. Furthermore, using multiple cameras with overlapping fields of view significantly increases the area that can be scanned in a single pass, reducing the number of scans needed and streamlining the mapping process.

Therefore, the use of a rotating device with RGB-D cameras offers significant advantages over handheld scanning for scene reconstruction in unknown indoor environments. The combination of controllable camera motion, efficient use of computational resources, extended field of view, effective loop closure, and improved data consistency makes this approach highly effective for these tasks. How to make use of the fixed scanning device for 3D Gaussian Splatting SLAM is introduced in the following sections.

\subsection{3D Gaussian Splatting}
Similar to popular GS-based approaches~\cite{kerbl3Dgaussians}, the 3D Gaussian ellipsoid representation is used as the only map primitive format in this paper. In this section, we briefly introduce the background and operations of 3D Gaussian Splatting. First, the representation of each 3D Gaussian $\mathcal{G}$ is denoted as 
\begin{equation}
    \mathcal{G}= [\bm{\mu} \; \mathbf{S} \; \mathbf{U} \; \mathbf{c} \; o]
\end{equation}
here $\bm{\mu}$, $\mathbf{c}$, $o$ are the mean vector, color, and opacity of $\mathcal{G}$, respectively. In our system, it is convenient to obtain the point clouds in camera coordinates based on the depth map and the intrinsic matrix, which can be transferred to the world coordinate frame via the rigid transformation $\mathbf{T}_{wc}$, from camera coordinates to world coordinates.
Given point clouds in the world coordinate frames, their positions are used to build the mean vectors of 3D Gaussians. Then, scaling matrix $\mathbf{S}$ and rotation $\mathbf{U}$ are to build the covariance matrix $\bm{\Sigma}$ of $\mathcal{G}$ via the following formulate:
\begin{equation}
    \bm{\Sigma}  = \mathbf{U}\mathbf{S}\mathbf{S}^T\mathbf{U}^T,
    \label{eq:sigma}
\end{equation}
here $\mathbf{S}$ can be represented as a diagonal matrix with $diag(s_x \; s_y \;s_z)$, and $\mathbf{U} = [  \mathbf{U}_x\; \mathbf{U}_y\; \mathbf{U}_z], \textbf{U} \in SO(3)$ is the orthornormal matrix. In this paper, to reduce the size of parameters. $s_x$, $s_y$, and $s_z$ are set as equal and the color $\mathbf{c}$ is using color of pixel directly.

After obtaining the 3D Gaussian representation $\mathcal{G}_{w}$ in the world coordinate frame, we continue to describe the view rendering process. To render images, 3D Gaussians in the world coordinates are projected onto 2D image planes based on transformation $\mathbf{T}_{cw}$. First, the splatting operation is to form 2D Gaussians ($\mathcal{N}(\bm{\mu}_I, \bm{\Sigma}_I )$) on image planes from 3D Gaussians ($\mathcal{N}(\bm{\mu}_w, \bm{\Sigma}_w )$) in the world coordinate based on camera poses. The process can be described as:
\begin{equation}
\left \{
\begin{split}
     \bm{\mu}_{I} &=\Pi(\mathbf{T}_{cw} \bm{\mu}_w)  \\
    \bm{\Sigma}_{I} &= \mathbf{J}\mathbf{W}_{cw}\bm{\Sigma}_{w}\mathbf{W}_{cw}^T\mathbf{J}^T
\end{split} \right.
\end{equation}
 here  $\mathbf{T}_{cw} = \left[\begin{array}{cc}
      \mathbf{W}_{cw}& \mathbf{t}_{cw}  \\
      \mathbf{0}& 1 
\end{array}\right], \mathbf{T}_{cw}\in SE$(3) is the camera pose from the world to the camera coordinates, $\mathbf{W}_{cw}$ and $\mathbf{t}_{cw}$ are rotational and translational components, respectively. $\mathbf{J}$ is the Jacobian matrix of the projective transformation. And the $\Pi(\cdot)$ represents the re-projection function.

Then, the alpha-blending operation is used to establish colors based on the following formulation:
\begin{equation}
    \mathcal{C}_{\mathbf{p}}=\sum_{i\in N} \mathbf{c}^io^i \prod_{j=1}^{i-1}(1-o^j).
\end{equation}
here $\mathbf{c}^i$ and $o^i$ represent the color and opacity of the $i^{th}$ Gaussian $\mathcal{G}^{i}$. Since the 3D Gaussian Splatting and blending operations are differentiable, Gaussian parameters can be trained gradually based on optimization solvers.

\section{Methodology}

\subsection{Tracking}
When three pairs of RGB-D images are fed to the system, as shown in Figure~\ref{fig:architecture}, we first make use of the motion model to estimate the initial pose of the current frame. Based on the initial pose, 3D Gaussian ellipsoids are used to render RGB and depth maps from this viewpoint. Then, we compute the residuals in photometric and geometric manners between rendered and observed images, where these residuals are used to build the loss function of a pair of RGB-D images as shown in Equation~\ref{eq:track_loss}. 
Similar to SplaTAM~\cite{keetha2024splatam}, we utilize the silhouette mask \( S(\mathbf{p}) \) to capture scene density and refine the initial camera pose by minimizing the joint loss function derived from color and depth maps:
\begin{equation}
    L_t = \sum_{\mathbf{p}} (S(\mathbf{p}) > 0.99 ) (L_1(E_{geo})+0.5L_1( E_{pho} ))
    \label{eq:track_loss}
\end{equation}
where $E_{geo}$ and $ E_{pho}$ are defined as geometric and photometric loss functions between rendered images and observers, respectively. They are handled by the $L_1$ loss. To make the tracking module robust, these pixels with depth values $0$ and $S(\mathbf{p}) < 0.99$ are ignored in the loss computation. To be specific, $E_{geo}$ and $ E_{pho}$ are computed by
\begin{equation}
    \left\{ 
    \begin{split}
        E_{pho} &= C(\mathbf{T}_{cw}, \mathcal{G}_w)-\Bar{C} \\
        E_{geo} &= D(\mathbf{T}_{cw}, \mathcal{G}_w) -\Bar{D}
    \end{split}
    \right.
\end{equation}
here $C(\mathbf{T}_{cw}, \mathcal{G})$ and $D(\mathbf{T}_{cw}, \mathcal{G})$ render the Gaussians $\mathcal{G}$ from $\mathbf{T}_{cw}$ to RGB and depth images, respectively. And $\Bar{C}$ and $\Bar{D}$ are images obtained from the device.

To make full use of the input of the three cameras, we not only use the middle camera to render the color image and depth map during the tracking process but also use the upper and lower cameras to render the color image and depth map to calculate the joint loss $L_{track}$:
\begin{equation}
   L_{track} = L_{t-up} + L_{t-mid} + L_{t-down}
   \label{eq:loss_track}
\end{equation}
where $L_{t-up}$, $L_{t-mid}$, and $L_{t-down}$ are using the same computation as $L_t$. Since the upper and lower cameras overlap with the middle camera, intra-frame constraints can be formed, making the camera parameter update more accurate and stable.

\subsection{Keyframe Selection and Gaussian Densification}\label{mapping}

To improve the tracking and rendering efficiency of the system, we save each $5^{th}$ frame as a keyframe, instead of using all input images to jointly optimize the Gaussians and camera poses. For each keyframe, we store color and depth images from all three cameras. In addition to these keyframes, we maintain a separate collection of general images as \textit{rand-list}, which are selected from the middle of adjacent keyframes. These general images only capture the color and depth information from the middle camera.

After finishing tracking the current frame, we perform a Gaussian densification operation to enhance the 3D Gaussian map based on the estimated camera pose. Unlike the densification process in 3DGS~\cite{kerbl3Dgaussians}, which focuses solely on existing map information, our approach integrates data from both general and keyframe images. The process begins by identifying which pixel areas are inadequately represented by the existing 3D Gaussians, using a densification detection mask~\cite{keetha2024splatam}. This mask highlights regions where the map's density is insufficient. For each pixel in these regions, new Gaussians are initialized to improve the map's overall density and accuracy.

Furthermore, in the densification process, the $j^{th}$ new Gaussian $\mathcal{G}^{j}$ is associated with the anchor frame $F_i$ by collecting timestamp $t_i$ to the Gaussian $_i\mathcal{G}^{j}$. Therefore, we add another parameter to each Gaussian, and the representation of $_i\mathcal{G}^{j}$ can be defined as 
\begin{equation}
    _i\mathcal{G}^{j} = [\bm{\mu} \; \bm{\Sigma} \; \mathbf{c} \; o \; t_i ]
\end{equation}
where $\bm{\Sigma}$ is computed via Equation~\ref{eq:sigma}.

Then, we implement the Gaussian parameter optimization stage by fixing the camera poses and refining the Gaussian parameters. Similar to the camera tracking stage, we use a loss function to optimize the Gaussian parameters. Referring to 3D Gaussian splatting~\cite{kerbl3Dgaussians}, the loss function for optimizing Gaussian parameters includes an additional D-SSIM~\cite{kerbl3Dgaussians} term compared to the tracking loss function:
\begin{equation}
    L_m = \sum_{\mathbf{p}}(0.8L_1(E_{pho})+0.2L_{D-SSIM}+2L_1(E_{geo})).
    \label{es:loss}
\end{equation}

Due to the input from the three cameras, our system needs to optimize more Gaussian parameters, making the optimization problem more complex. Therefore, similar to the tracking section, we use the loss function Equation~\ref{es:loss} from multiple viewpoints to jointly optimize the Gaussian parameters. Specifically, we use the current frame and randomly select a keyframe from the keyframe database that has visual overlaps with the current frame to calculate the loss:
\begin{equation}
    L_{map} = \lambda L_{m-up} + L_{m-mid} + \lambda L_{m-down}
\end{equation}
here $ L_{m-up}$, $L_{m-mid}$, and $L_{m-down}$ are using the same computation manner as $L_m$.

Additionally, to prevent forgetting the global map, in the optimization process, we randomly select a frame from the keyframe list and a frame from \textit{rand-list} that do not overlap with the current frame. We calculate the loss using the middle camera and add this to the loss function $L_{map}$ used for optimizing the Gaussian parameters. If both novel and history Gaussians are captured in these frames, we calculate the loss using only the novel Gaussians, and the method to obtain different types of Gaussians is introduced in Section~\ref{Loop Closure}.

\subsection{Loop Closure}\label{Loop Closure}
In the camera pose tracking process, the drift is typically witnessed, which leads to the noisy Gaussian map and worse initials for Gaussian Splatting optimization. To solve the problem, we propose a novel loop closure module for Gaussian Splatting SLAM, which contains three stages: loop detection, drift estimation, and optimization. 

\paragraph{Loop Detection}~\label{sec:loop_detection}
During tracking and Gaussian parameter optimization, 3D Gaussians are classified into two groups, history and novel Gaussian sets, according to the timestamps collected in Gaussian parameters. For example, if the difference between Gaussian's timestamp and the timestamp $t_c$ of the current frame exceeds a fixed threshold $\tau_t$, the 3D Gaussian $_i\mathcal{G}^j$ will be added into the set of $\mathcal{S}_h^c$, otherwise will be added into $\mathcal{S}_n^c$, 
\begin{equation}
\left\{
\begin{split}
    \mathcal{S}_h^c &= [_i\mathcal{G}^j | \; \tau_t \leq |t_i-t_c| ]  \\
    \mathcal{S}_n^c &= [_i\mathcal{G}^j | \; \tau_t > |t_i-t_c| ] 
\end{split} \right.
\end{equation}
here $\mathcal{S}_h^c$ and $\mathcal{S}_n^c$ are the history and novel Gaussian sets of frame $F_c$, respectively. It has to be mentioned that both history and novel sets of Gaussians for each frame are collected virtually instead of saving them in memories to improve efficiency.

Since newly generated Gaussians in \(\mathcal{S}_n^c\) are unaffected by noise, only Gaussians in \(\mathcal{S}_n^c\) are involved in tracking and optimizing Gaussian parameters for the current frame. Once the camera pose \(\mathbf{T}_{cw}\) is obtained, we re-project the positions of Gaussians from both the history set \(\mathcal{S}_h^c\) and the novel set \(\mathcal{S}_n^c\) onto the image plane. 
We then count the number of these re-projected positions on the image plane, denoting them as \(N_n^c\) for the novel set and \(N_h^c\) for the history one. If the ratio \(\gamma = \frac{N_h^c}{N_n^c}\) exceeds a predefined threshold \(\tau_r\), we can know that the scenario is already detected before, then a candidate loop frame is collected by the system.

Additionally, another strategy is proposed to obtain the best one from candidates by rendering two images at this viewpoint based on two sets $\mathcal{S}_n^c$ and $\mathcal{S}_h^c$, respectively. Based on the SSIM metric, we continue to compare the SSIM distance between these two images. We then select the one that has the best SSIM similarity as the final loop frame.

\paragraph{Pose Graph Optimization}\label{pose graph}
Given the loop frame detected in the former stage, a lightweight pose graph is implemented in this paper to remove camera pose drift and achieve accurate trajectories. In our pose graph, each vertex is built based $\mathbf{T}_{c_iw}$ and there are two types of edges $\mathcal{E}$ to connect these vertices, where the first one is estimated based on the relative transformations of adjacent camera poses, and the second component is the accurate relative pose \( \mathbf{T}_{c_ic_j} \) of the frames at endpoints of the loop, which can be determined by rendering history Gaussians to the current frame $F_j$ using Equation~\ref{eq:loss_track}. By analyzing the timestamps of the historical Gaussian ellipsoids, we identify the dominant class \( t_i \) that has the highest count as another endpoint of the loop, $F_i$. Therefore, the pose graph can be defined as
\begin{equation}
    \begin{split}
        \min_{\mathbf{W}_{wc_i},\mathbf{t}_{wc_i}} \sum_{(i,j)\in\mathcal{E}} 
        &( L_{\mathbb{R}^3} (\mathbf{t}_{ij}, \mathbf{W}_{wc_i}^\top(\mathbf{t}_{wc_j} - \mathbf{t}_{wc_i}))^2 \\
        & + L_{SO(3)}(\mathbf{W}_{ij}, \mathbf{W}_{wc_i}^\top\mathbf{W}_{wc_j})^2 ) \\
    \end{split}
    \label{eq:pose_graph}
\end{equation}
where $L_{\mathbb{R}^3}(\mathbf{t}_a, \mathbf{t}_b)$ denotes the Euclidean distance between two translation vectors, while $L_{SO(3)}(\mathbf{W}_a, \mathbf{W}_b )$ ) represents a distance metric between two orientations in SO(3). In this way, the pose graph optimization module allows us to adjust the camera poses of all frames. 

\paragraph{Gaussian Updating and Bundle Adjusting}
Benefiting from the proposed Gaussian-frame association strategy, it is easy to correct errors in 3D Gaussian maps via a linear updating process.  
For example, for a 3D Gaussian with timestamp $t_i$, its position $\bm{\mu}$ will be updated according to the refined pose of the anchor frame via the following function
\begin{equation}
    \bm{\mu}^* = \mathbf{T}_{opt}\mathbf{T}^{-1}_{origin}\bm{\mu}
\end{equation}
where $\bm{\mu}^*$ is the updated position of $\mathcal{G}$, and $\mathbf{T}^{-1}_{origin}$ represents the camera poses before pose graph optimization, and $\mathbf{T}_{opt}$ represents the camera poses after pose graph optimization. Benefiting from the Gaussian representation, the covariance matrix $\bm{\Sigma}$ does not need to be updated, which reduces computation burdens of the loop closure module.

After updating camera poses based on pose graph refinement and the positions of Gaussians, we then adjust the camera poses of all frames to achieve more accurate and consistent camera poses aligned with the 3D Gaussian map. 
Compared to the lightweight pose graph optimization before, this step refines both keyframes and frames in \textit{rand-list} based on photometric and geometric constraints. Specifically, we first mix frames in the keyframe database and \textit{rand-list} together, and then divide these frames into several bins, where each bin has $N$ frames adjacent in the timestamp. What is more, we optimize camera poses collected in each bin $\mathcal{B}_j$ via an efficient bundle adjustment, in Equation~\ref{eq:local_ba}, that only considers the image of the middle camera,
\begin{equation}
    L_{local} = \sum_{i \in \mathcal{B}_j}L_{t-mid, i}.
    \label{eq:local_ba}
\end{equation}

Once the camera poses for the keyframes and rand-list frames are optimized, we fix these poses and perform pose graph optimization following the graph structure estimated in Equation~\ref{eq:pose_graph}, to refine the poses of all non-keyframes and non-rand-list frames. This approach ensures that the camera poses across the entire sequence are accurately adjusted, improving the overall consistency of the 3D Gaussian map and trajectories.

% 无抖动虚拟数据集PSNR、SSIM、LPIPS、Depth L1指标对比
\begin{table*}[ht]
    \centering
    \captionsetup[table*]{singlelinecheck=off}
	\renewcommand{\arraystretch}{1.2} %rows, default value is 1.0
	\setlength{\tabcolsep}{9pt}
    \resizebox{\textwidth}{!}{
        \begin{tabular}{c|c|ccccccc}
        \toprule
        Method                  & Metric            & \multicolumn{1}{l}{room ($2^{\circ}$) } & \multicolumn{1}{l}{room ($3^{\circ}$)} & \multicolumn{1}{l}{room ($4^{\circ}$)} & \multicolumn{1}{l}{office($2^{\circ}$)} & \multicolumn{1}{l}{office($3^{\circ}$)} & \multicolumn{1}{l}{office ($4^{\circ}$)} & \multicolumn{1}{l}{Avg.} \\
        \hline
        \multirow{4}{*}{MonoGS\cite{Matsuki:Murai:etal:CVPR2024}} & 
        PSNR{[}dB{]} $\uparrow$  & 30.09  & 27.21  & 25.84  & 30.71   & 29.73   & 27.22  & 28.467  \\
        & SSIM$\uparrow$   & 0.914   & 0.886   & 0.864  & 0.859   & 0.855   & 0.827  & 0.868       \\
        & LPIPS$\downarrow$  & 0.166  & 0.22  & 0.275    & 0.185   & 0.21    & 0.263   & 0.220   \\
        & Depth L1{[}cm{]}$\downarrow$  & 1.3    & 2.3      & 2.9     & 2.9   & 2.7  & 4.4  & 2.75 \\
        \hline
        \multirow{4}{*}{Gaussian-SLAM\cite{yugay2023gaussianslam}} & 
        PSNR{[}dB{]} $\uparrow$  & 33.55  & 33.55  & 32.22  & 33.01   & 34.17   & 34.97  & 33.578  \\
        & SSIM$\uparrow$   & 0.965   & 0.963   & 0.953 & 0.962   & 0.964   & 0.972  & 0.963       \\
        & LPIPS$\downarrow$  & 0.108  & 0.117  & 0.142   & 0.098   & 0.084    & 0.074   & 0.104   \\
        & Depth L1{[}cm{]}$\downarrow$  & 0.71    & 0.80      &0.75      & 1.1   & 1.2  & 1.1  & 0.977 \\
        \hline
        \multirow{4}{*}{SplaTAM\cite{keetha2024splatam}} & 
        PSNR{[}dB{]} $\uparrow$  & 34.88  & 34.90  & 35.45  & 34.10   & 33.81   & 33.79  & 34.473  \\
        & SSIM$\uparrow$   & 0.973   & 0.973   & 0.978 & 0.970   & 0.972   & 0.971  & 0.973       \\
        & LPIPS$\downarrow$  & 0.090  & 0.082  & 0.055    & 0.097   & 0.089    & 0.081   & 0.082   \\
        & Depth L1{[}cm{]}$\downarrow$  & 0.60    & 0.61      & 0.55     & 0.82   & 0.80  & 0.82  & 0.700 \\
        \hline
        \multirow{4}{*}{Ours} & 
        PSNR{[}dB{]} $\uparrow$  & \textbf{38.19}  & \textbf{38.03}  & \textbf{37.55}  & \textbf{37.39}   & \textbf{37.11}   & \textbf{37.11}  & \textbf{37.565}  \\
        & SSIM$\uparrow$   & \textbf{0.993}   & \textbf{0.993}   & \textbf{0.992} & \textbf{0.988}   & \textbf{0.988}   & \textbf{0.987}  & \textbf{0.990}       \\
        & LPIPS$\downarrow$  & \textbf{0.035}  & \textbf{0.033}  & \textbf{0.039}    & \textbf{0.032}   & \textbf{0.032}    & \textbf{0.033}   & \textbf{0.034}   \\
        & Depth L1{[}cm{]}$\downarrow$  & \textbf{0.40}    & \textbf{0.38}   & \textbf{0.39}     & \textbf{0.47}   & \textbf{0.49}  & \textbf{0.49}  & \textbf{0.437} \\
        \bottomrule
        \end{tabular}}
\caption{\textbf{Rendering and reconstruction performance on virtual datasets without noise and jitters.} Results with best accuracy are highlighted by \textbf{bold} font. }
\label{virtual w/o jitters}
\end{table*}

% 有抖动虚拟数据集PSNR、SSIM、LPIPS、Depth L1指标对比
\begin{table*}[ht]
    \centering
    \captionsetup[table*]{singlelinecheck=off}
	\renewcommand{\arraystretch}{1.2} %rows, default value is 1.0
	\setlength{\tabcolsep}{7pt}
    \resizebox{\textwidth}{!}{
        \begin{tabular}{c|c|ccccccc}
        \toprule
        Method                  & Metric            & \multicolumn{1}{l}{room2n1} & \multicolumn{1}{l}{room3n1} & \multicolumn{1}{l}{room4n1} & \multicolumn{1}{l}{office2n1} & \multicolumn{1}{l}{office3n1} & \multicolumn{1}{l}{office4n1} & \multicolumn{1}{l}{Avg.} \\
        \hline
        \multirow{4}{*}{MonoGS\cite{Matsuki:Murai:etal:CVPR2024}} & 
        PSNR{[}dB{]} $\uparrow$  & 30.25  & 27.82  & 25.61  & 30.91   & 27.75   & 28.19  & 28.422  \\
        & SSIM$\uparrow$   & 0.916   & 0.893   & 0.863  & 0.863   & 0.834   & 0.839  & 0.868       \\
        & LPIPS$\downarrow$  & 0.165  & 0.21  & 0.267    & 0.170   & 0.249    & 0.254   & 0.219   \\
        & Depth L1{[}cm{]}$\downarrow$  & 1.4    & 2.5     & 3.7     & 2.9   & 4.2  & 4.6  & 3.217 \\
        \hline
        \multirow{4}{*}{Gaussian-SLAM\cite{yugay2023gaussianslam}} & 
        PSNR{[}dB{]} $\uparrow$  & 35.47  & 35.11  & 34.69  & 35.48   & 35.05   & 35.24  & 35.173  \\
        & SSIM$\uparrow$   & 0.980   & 0.978   & 0.976 & 0.986   & 0.980   & 0.980  & 0.980       \\
        & LPIPS$\downarrow$  & 0.093  & 0.094  & 0.094    & 0.068   & 0.071    & 0.071   & 0.081   \\
        & Depth L1{[}cm{]}$\downarrow$  & 0.56   & 0.57 & 0.58  & 0.91   & 0.86  & 0.94  & 0.736 \\
        \hline
        \multirow{4}{*}{SplaTAM\cite{keetha2024splatam}} & 
        PSNR{[}dB{]} $\uparrow$  & 35.90  & 36.10  & 36.51  & 34.49   & 34.39   & 34.82  & 35.368  \\
        & SSIM$\uparrow$   & 0.979   & 0.981   & 0.981 & 0.975   & 0.974   & 0.997  & 0.978       \\
        & LPIPS$\downarrow$  & 0.081  & 0.069  & 0.063    & 0.084   & 0.075    & 0.071   & 0.074   \\
        & Depth L1{[}cm{]}$\downarrow$  & 0.57    & 0.65      & 0.68     & 0.88   & 0.94  & 1.03  & 0.792 \\
        \hline
        \multirow{4}{*}{Ours} & 
        PSNR{[}dB{]} $\uparrow$  & \textbf{39.43}  & \textbf{39.39}  & \textbf{39.44}  & \textbf{37.91}   & \textbf{38.03}   & \textbf{37.87}  & \textbf{38.678}  \\
        & SSIM$\uparrow$   & \textbf{0.994}   &\textbf{ 0.994}   & \textbf{0.994} & \textbf{0.990}   & \textbf{0.990}   & \textbf{0.990}  & \textbf{0.992}       \\
        & LPIPS$\downarrow$  & \textbf{0.026}  &\textbf{ 0.026}  & \textbf{0.027}    & \textbf{0.027}   & \textbf{0.027}    & \textbf{0.028}   & \textbf{0.026}   \\
        & Depth L1{[}cm{]}$\downarrow$  & \textbf{0.49}    & \textbf{0.49}   & \textbf{0.49}     & \textbf{0.68}   & \textbf{0.68}  & \textbf{0.69}  & \textbf{0.586} \\
        \bottomrule
        \end{tabular}}
\caption{\textbf{Rendering and reconstruction performance on virtual datasets with noise and jitters.} Results with best accuracy are highlighted by \textbf{bold} font.}
\label{virtual}
\end{table*}

\section{Experiment}
% implementation

\subsection{Implementation Details}
%In this section, we evaluate the performance of the proposed method through extensive experiments compared to state-of-the-art Gaussian Splatting methods. 

Our method is implemented in Python using the PyTorch framework, incorporating CUDA code for time-critical rasterization and gradient computation of Gaussian splatting, and we run our SLAM on a desktop with Intel(R) Xeon(R) Silver 4210R and a single NVIDIA GeForce RTX 3080 Ti. In all experiments, we consider every fifth frame as a keyframe.
Furthermore, we set the thresholds $\tau_r$=0.8, $\tau_{ssim}$=25, \(\tau_t\)=60 for all experiments and use $\lambda$=0.5 for virtual datasets, $\lambda$=0.4 for real-world datasets.  
\subsection{Datasets}~\label{Datasets}
There are synthetic~\cite{10582466} and real-world~\cite{10582466} datasets used in the evaluation section. We introduce the details of these datasets in this section.

\textit{Virtual Dataset.}
The synthetic dataset is derived from the ICL-NUIM dataset~\cite{6907054}. It simulates the motion of three cameras during device rotation, rendering RGB-D images separately for both `living room' and `office' scenes from the ICL-NUIM dataset. The images are captured at 30 Hz with a resolution of $640 \times 480$ pixels.The terms `room' and `office' in the dataset names denote different scene categories. The numbers $2$, $3$, $4$ following `room' or `office' indicate that the motor rotates by $2$, $3$, $4$ degrees per frame. The suffix `n1' signifies that random perturbations were applied to the motor's rotation axis coordinates, simulating the unavoidable shake motion observed during real-world data acquisition, and the noise was added to the depth images to simulate real-world conditions.

% 虚拟数据集渲染结果
\begin{figure*}[htbp]
	\centering
    \captionsetup[subfloat]{labelformat=empty}%
	\subfloat{%
        \hspace{-5mm}%
        \rotatebox{90}{\scriptsize{~~~~~~~~~~~~~~~~~~~~~~~~\textbf{office3n1}}}
		\begin{minipage}[b]{0.25\linewidth}
			\centering
			\includegraphics[width=1\linewidth]{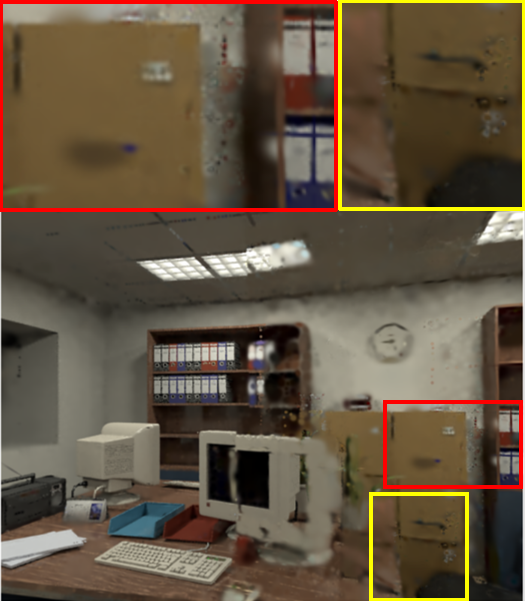}%
		\end{minipage}
	}%
    %\hspace{-2mm}
	\subfloat{%
		\begin{minipage}[b]{0.25\linewidth}
			\centering
			\includegraphics[width=1\linewidth]{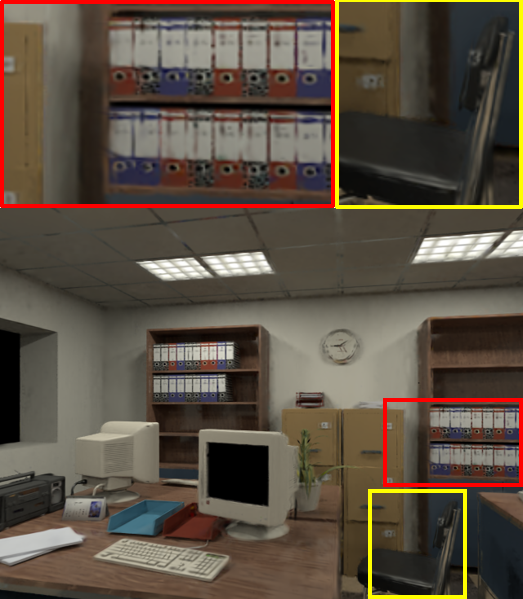}%
		\end{minipage}
	}%
    %\hspace{-2mm}
	\subfloat{%
		\begin{minipage}[b]{0.25\linewidth}
			\centering
			\includegraphics[width=1\linewidth]{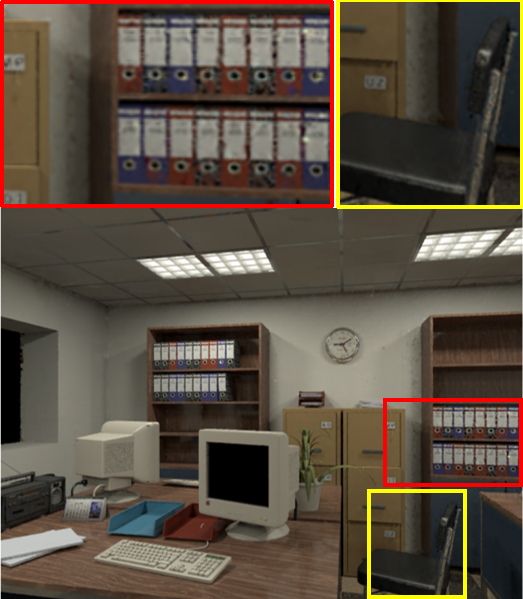}%
		\end{minipage}
	}%
    %\hspace{-2mm}
    \subfloat{%
		\begin{minipage}[b]{0.25\linewidth}
			\centering
			\includegraphics[width=1\linewidth]{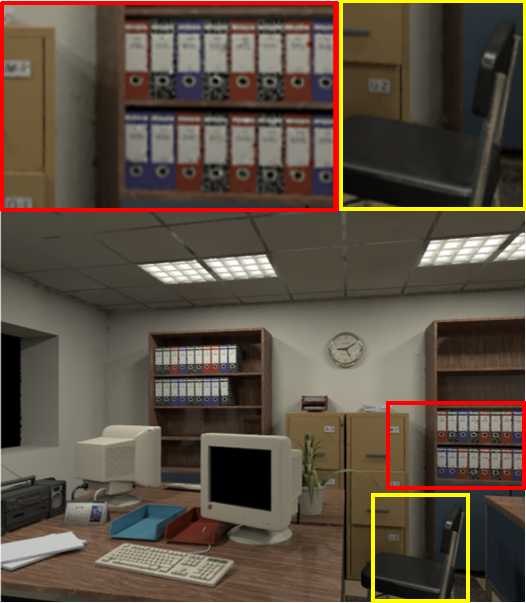}%
		\end{minipage}
	}\\
     % 第二行
	\vspace{-2mm}
    \subfloat{%
        \hspace{-5mm}%
        \rotatebox{90}{\scriptsize{~~~~~~~~~~~~~~~~~~~~~~~~~~~~~~~~~~\textbf{office4n1}}}
		\begin{minipage}[b]{0.25\linewidth}
			\centering
			\includegraphics[width=1\linewidth]{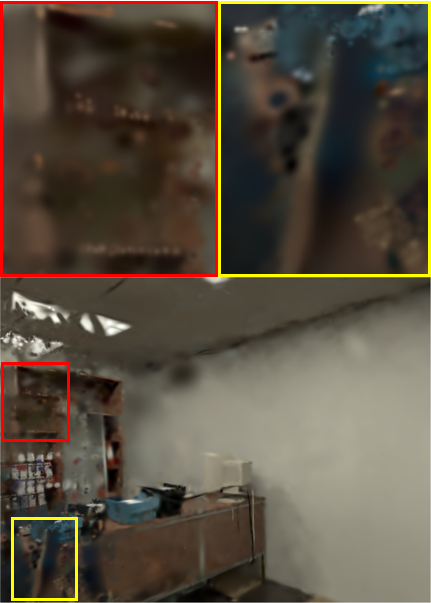}%
		\end{minipage}
	}%
    %\hspace{-4mm}
	\subfloat{%
		\begin{minipage}[b]{0.25\linewidth}
			\centering
			\includegraphics[width=1\linewidth]{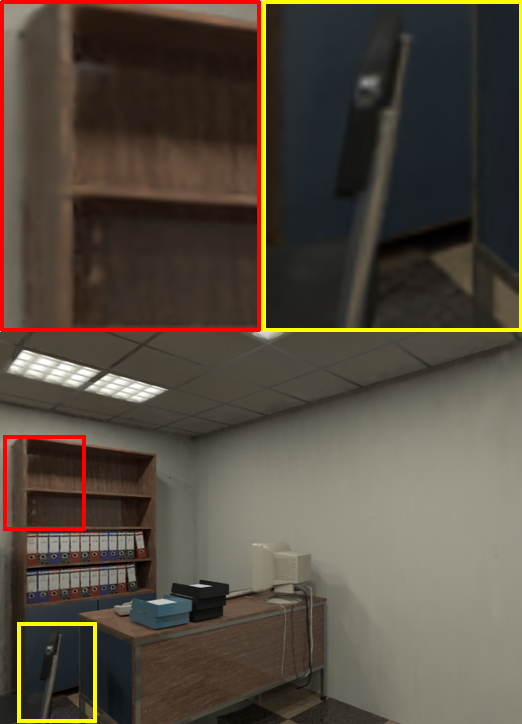}%
		\end{minipage}
	}%
    %\hspace{-4mm}
	\subfloat{%
		\begin{minipage}[b]{0.25\linewidth}
			\centering
			\includegraphics[width=1\linewidth]{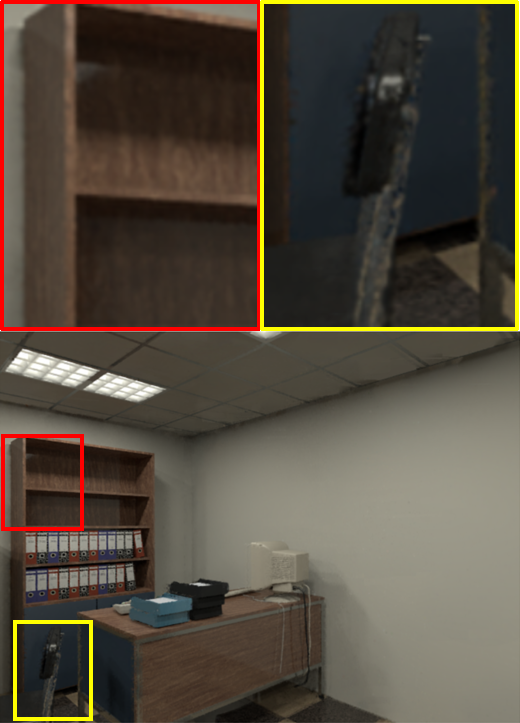}%
		\end{minipage}
	}%
    %\hspace{-4mm}
    \subfloat{%
		\begin{minipage}[b]{0.25\linewidth}
			\centering
			\includegraphics[width=1\linewidth]{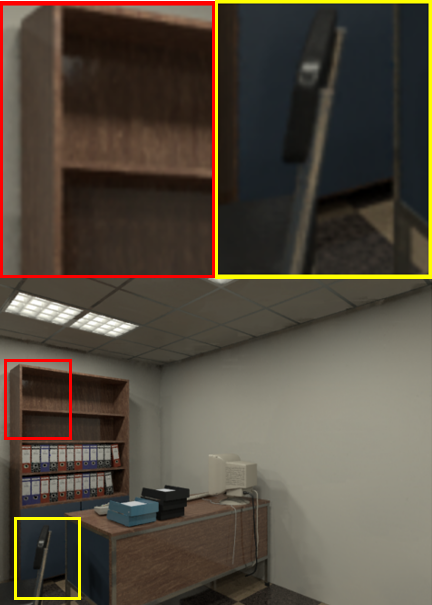}%
		\end{minipage}
	}\\
	\vspace{-2mm}%
    % 第三行
 %    \subfloat{%
 %        \hspace{-5mm}%
 %        \rotatebox{90}{\scriptsize{~~~~~~~~~~~~~~\textbf{room2n1}}}
	% 	\begin{minipage}[b]{0.25\linewidth}
	% 		\centering
	% 		\includegraphics[width=1\linewidth]{img/visual_result/room2n1/MonoGS.png}%
	% 	\end{minipage}
	% }%
 %    %\hspace{-4mm}
	% \subfloat{%
	% 	\begin{minipage}[b]{0.25\linewidth}
	% 		\centering
	% 		\includegraphics[width=1\linewidth]{img/visual_result/room2n1/Gaussian-SLAM.png}%
	% 	\end{minipage}
	% }%
 %    %\hspace{-4mm}
	% \subfloat{%
	% 	\begin{minipage}[b]{0.25\linewidth}
	% 		\centering
	% 		\includegraphics[width=1\linewidth]{img/visual_result/room2n1/splatam.png}%
	% 	\end{minipage}
	% }%
 %    %\hspace{-4mm}
 %    \subfloat{%
	% 	\begin{minipage}[b]{0.25\linewidth}
	% 		\centering
	% 		\includegraphics[width=1\linewidth]{img/visual_result/room2n1/ours.png}%
	% 	\end{minipage}
	% }\\
	% \vspace{-2mm}
	% %\setcounter{subfloat}{0}
    % 第四行
    \subfloat[MonoGS]{%
        \hspace{-5mm}%
		\rotatebox{90}{\scriptsize{~~~~~~~~~~~~~~~~~~~~~~\textbf{room3n1}}}
		\begin{minipage}[t]{0.25\linewidth}
			\centering
			\includegraphics[width=1\linewidth]{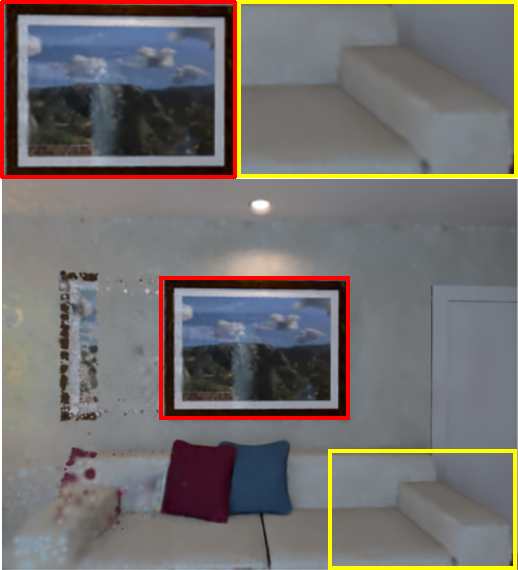}%
		\end{minipage}
	}%
	\subfloat[Gaussian-SLAM]{%
		\begin{minipage}[t]{0.25\linewidth}
			\centering
			\includegraphics[width=1\linewidth]{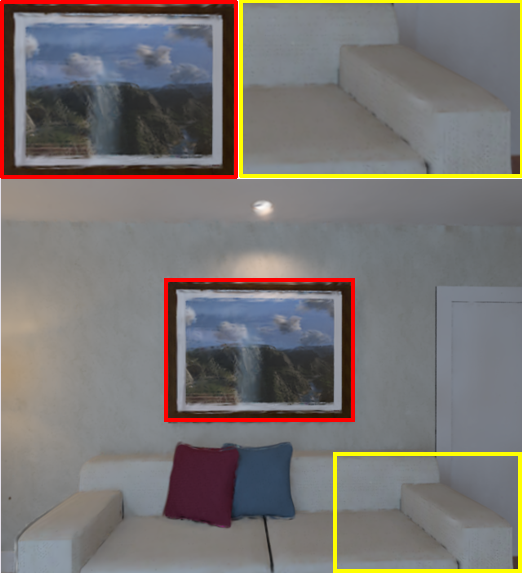}%
		\end{minipage}
	}%
	\subfloat[SplaTAM]{%
		\begin{minipage}[t]{0.25\linewidth}
			\centering
			\includegraphics[width=1\linewidth]{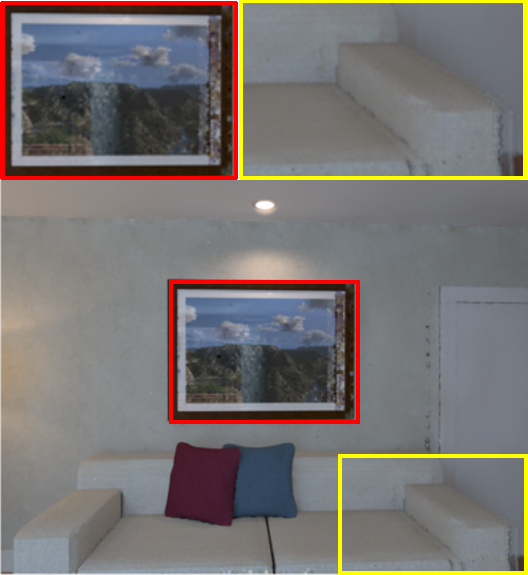}%
		\end{minipage}
	}%
    \subfloat[Ours]{%
		\begin{minipage}[t]{0.25\linewidth}
			\centering
			\includegraphics[width=1\linewidth]{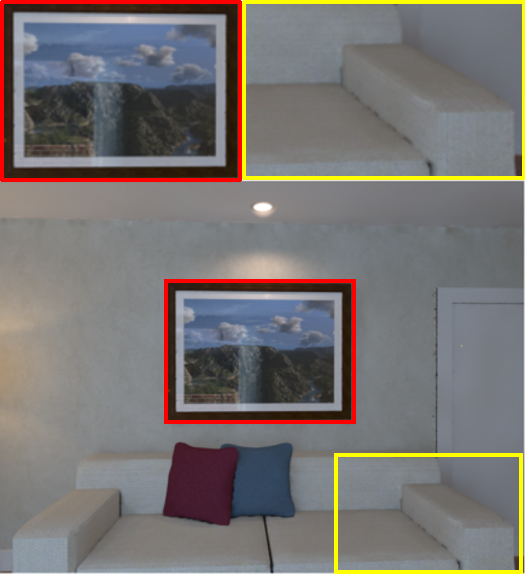}%
		\end{minipage}
	}
	\caption{\textbf{Comparison of novel view rendering in virtual sequences.} This is also supported by the quantitative results
    in Table~\ref{virtual w/o jitters} and \ref{virtual}.}
	\label{visual render}%
\end{figure*}

% 真实深度图和渲染深度图之差结果
\begin{figure*}[htbp]
	\centering
    \captionsetup[subfloat]{labelformat=empty}%
	\subfloat{%
        \hspace{-5mm}%
        \rotatebox{90}{\scriptsize{~~~~~~~~~~~~~~\textbf{office4n1}}}
		\begin{minipage}[b]{0.25\linewidth}
			\centering
			\includegraphics[width=1\linewidth]{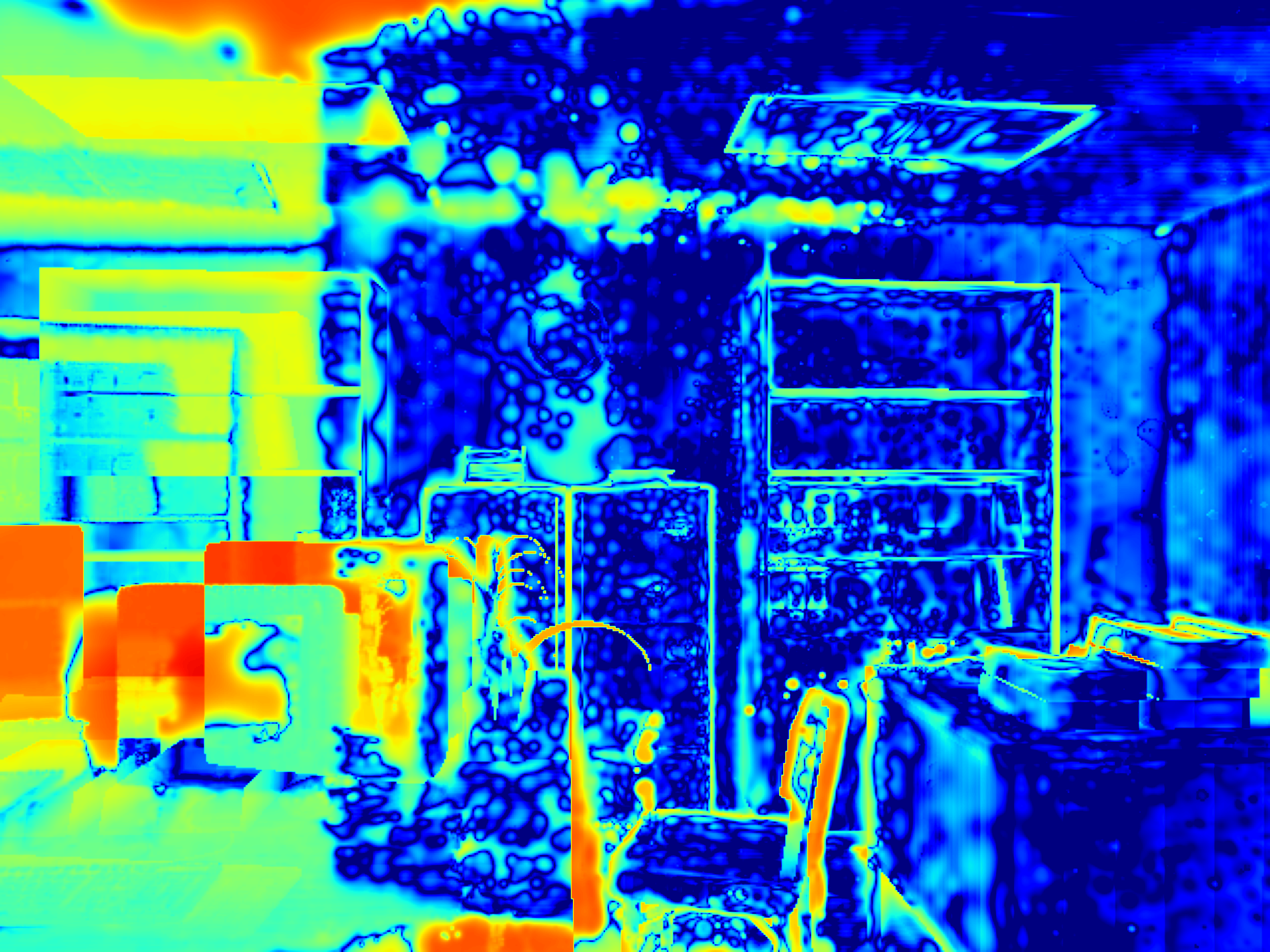}%
		\end{minipage}
	}%
    %\hspace{-2mm}
	\subfloat{%
		\begin{minipage}[b]{0.25\linewidth}
			\centering
			\includegraphics[width=1\linewidth]{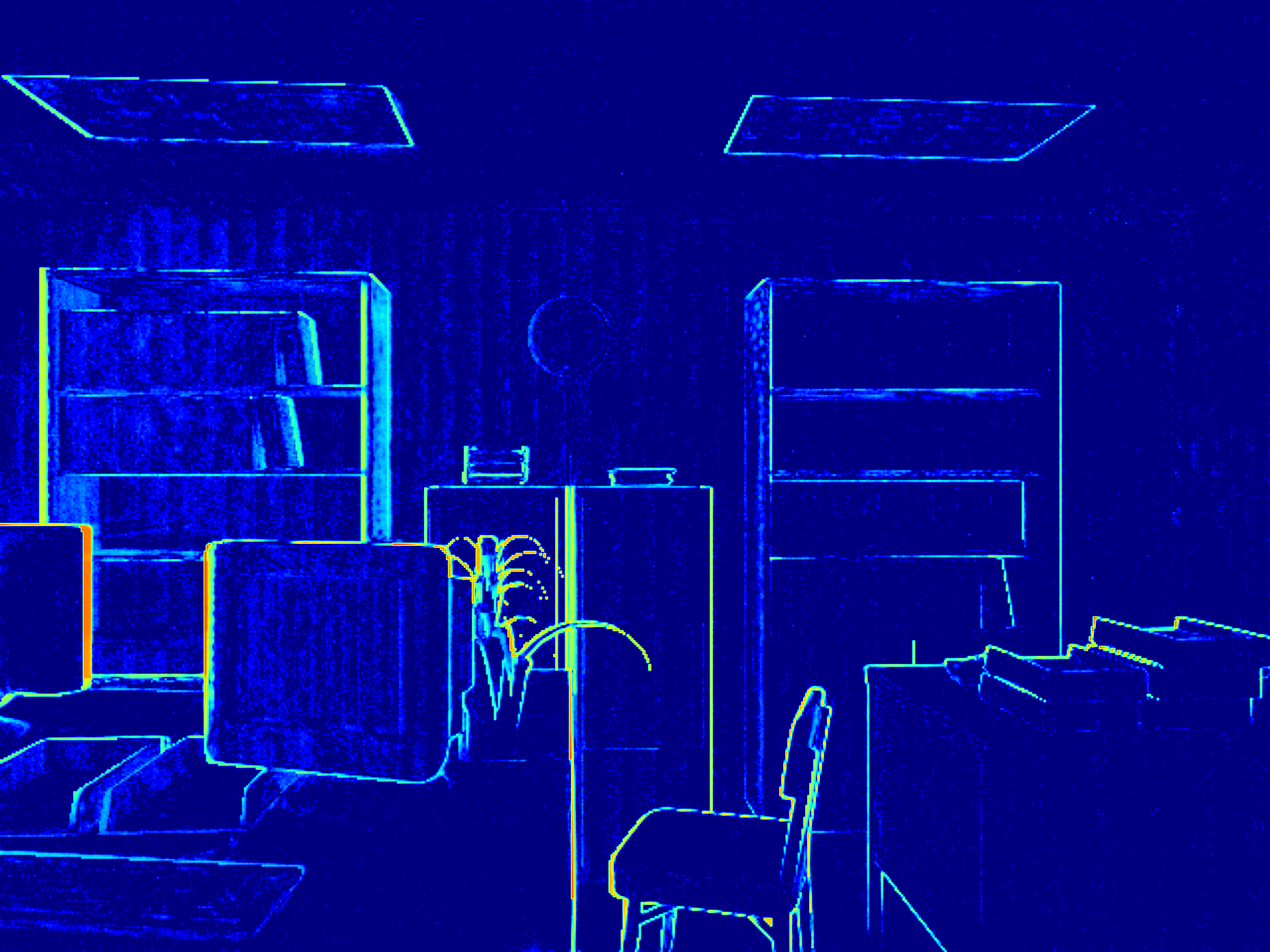}%
		\end{minipage}
	}%
    %\hspace{-2mm}
	\subfloat{%
		\begin{minipage}[b]{0.25\linewidth}
			\centering
			\includegraphics[width=1\linewidth]{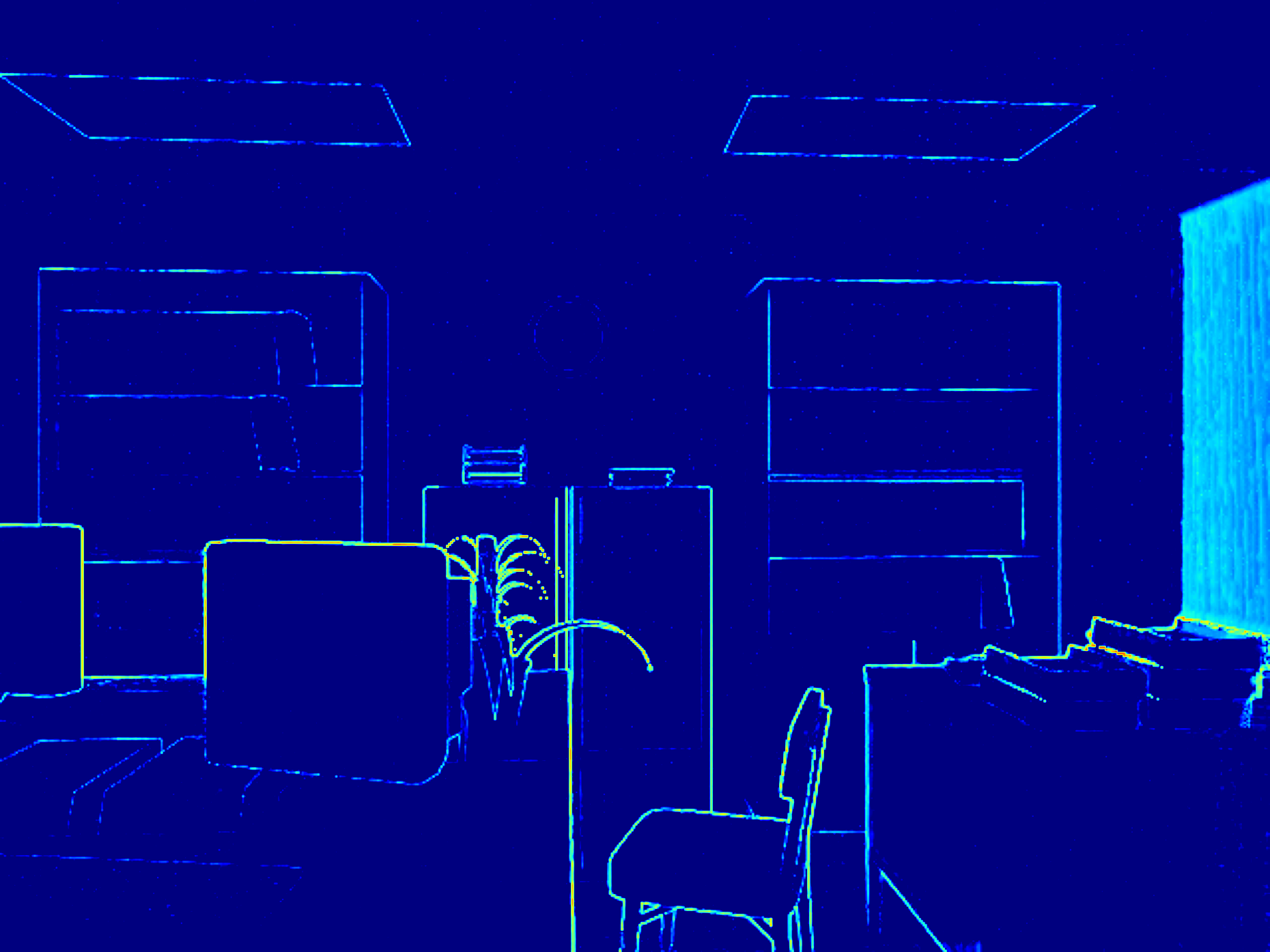}%
		\end{minipage}
	}%
    %\hspace{-2mm}
    \subfloat{%
		\begin{minipage}[b]{0.25\linewidth}
			\centering
			\includegraphics[width=1\linewidth]{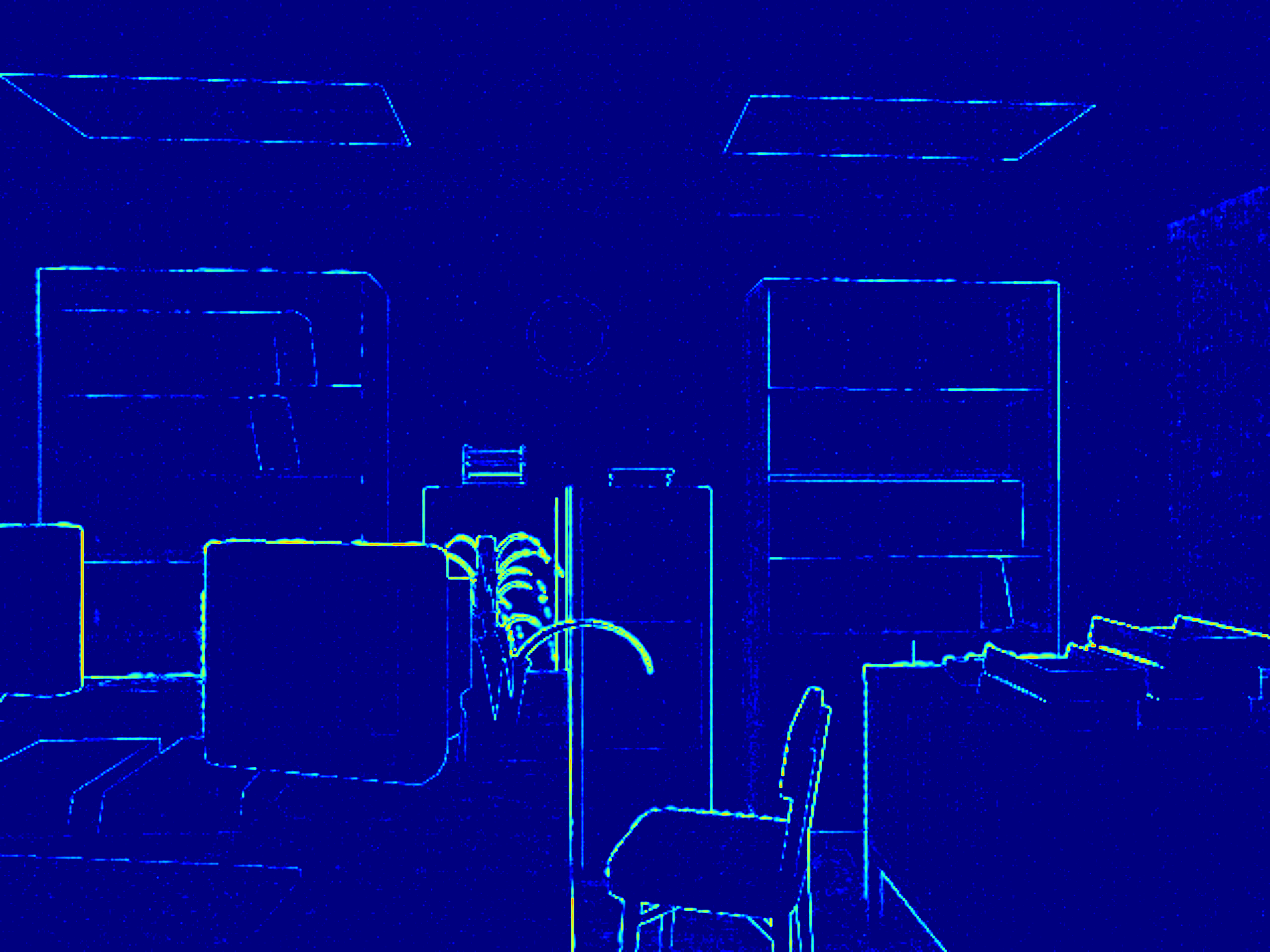}%
		\end{minipage}
	}\\
     % 第2行
	\vspace{-2mm}
    \subfloat{%
        \hspace{-5mm}%
        \rotatebox{90}{\scriptsize{~~~~~~~~~~~~~~\textbf{room4n1}}}
		\begin{minipage}[b]{0.25\linewidth}
			\centering
			\includegraphics[width=1\linewidth]{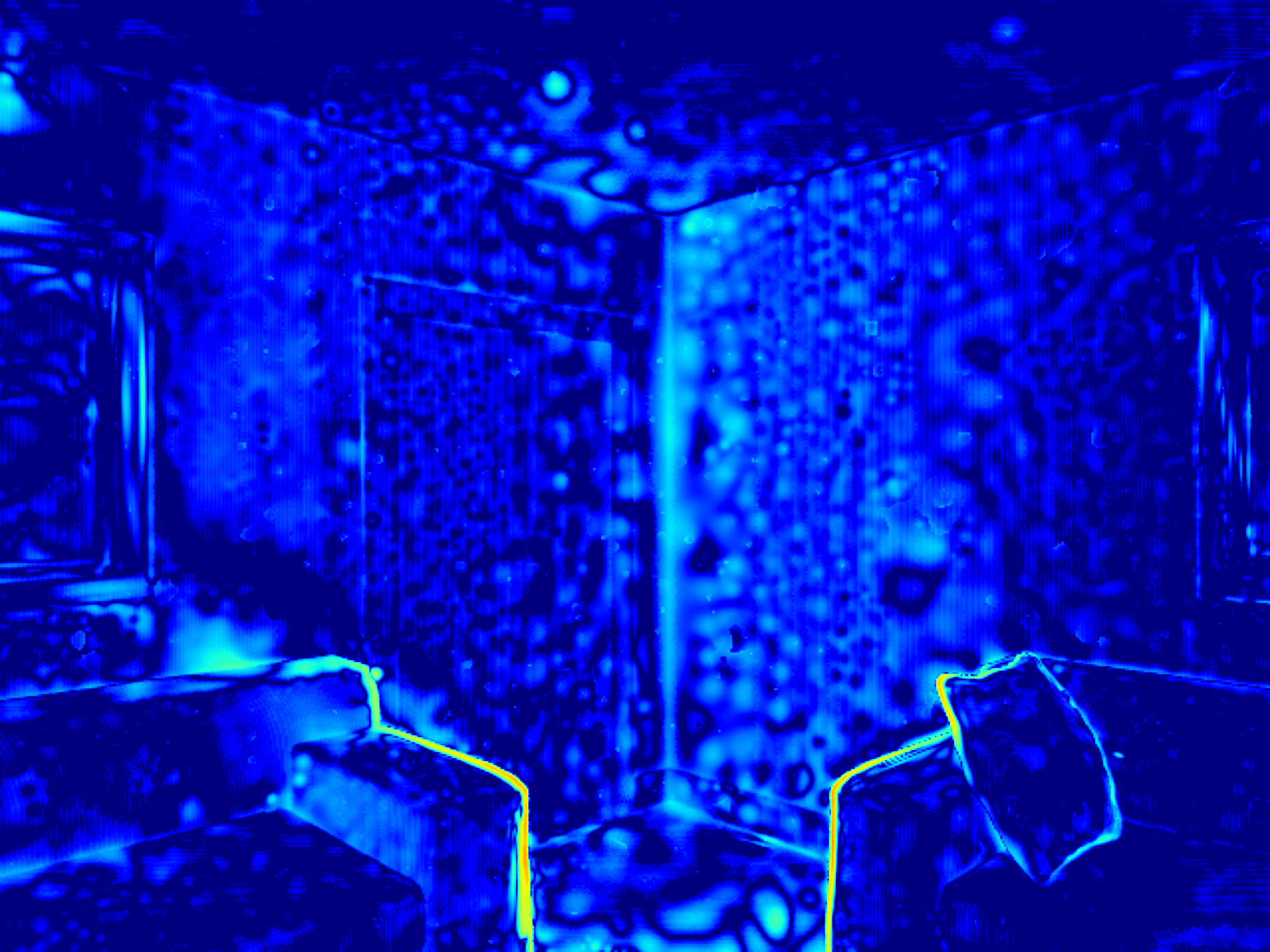}%
		\end{minipage}
	}%
    %\hspace{-4mm}
	\subfloat{%
		\begin{minipage}[b]{0.25\linewidth}
			\centering
			\includegraphics[width=1\linewidth]{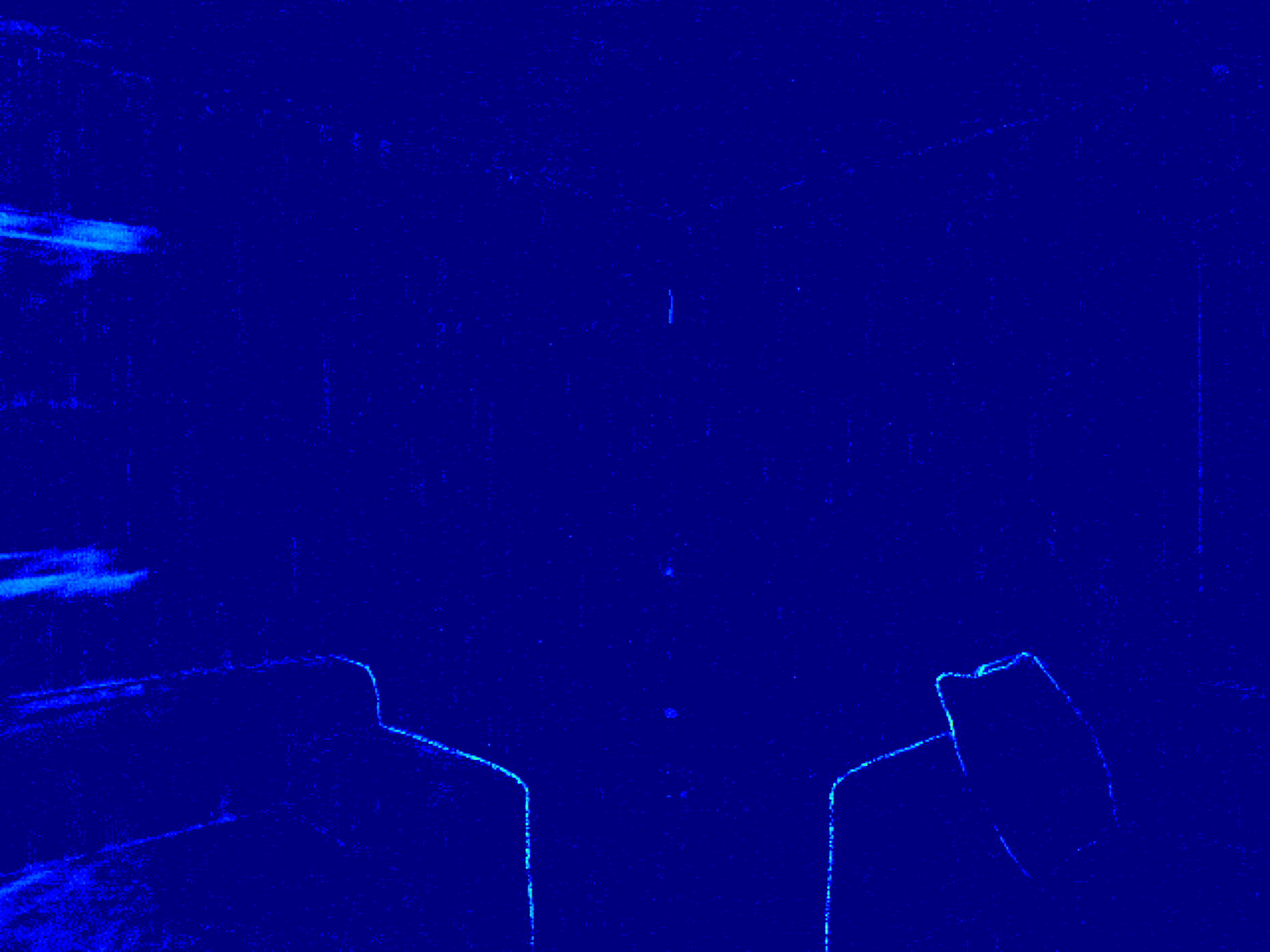}%
		\end{minipage}
	}%
    %\hspace{-4mm}
	\subfloat{%
		\begin{minipage}[b]{0.25\linewidth}
			\centering
			\includegraphics[width=1\linewidth]{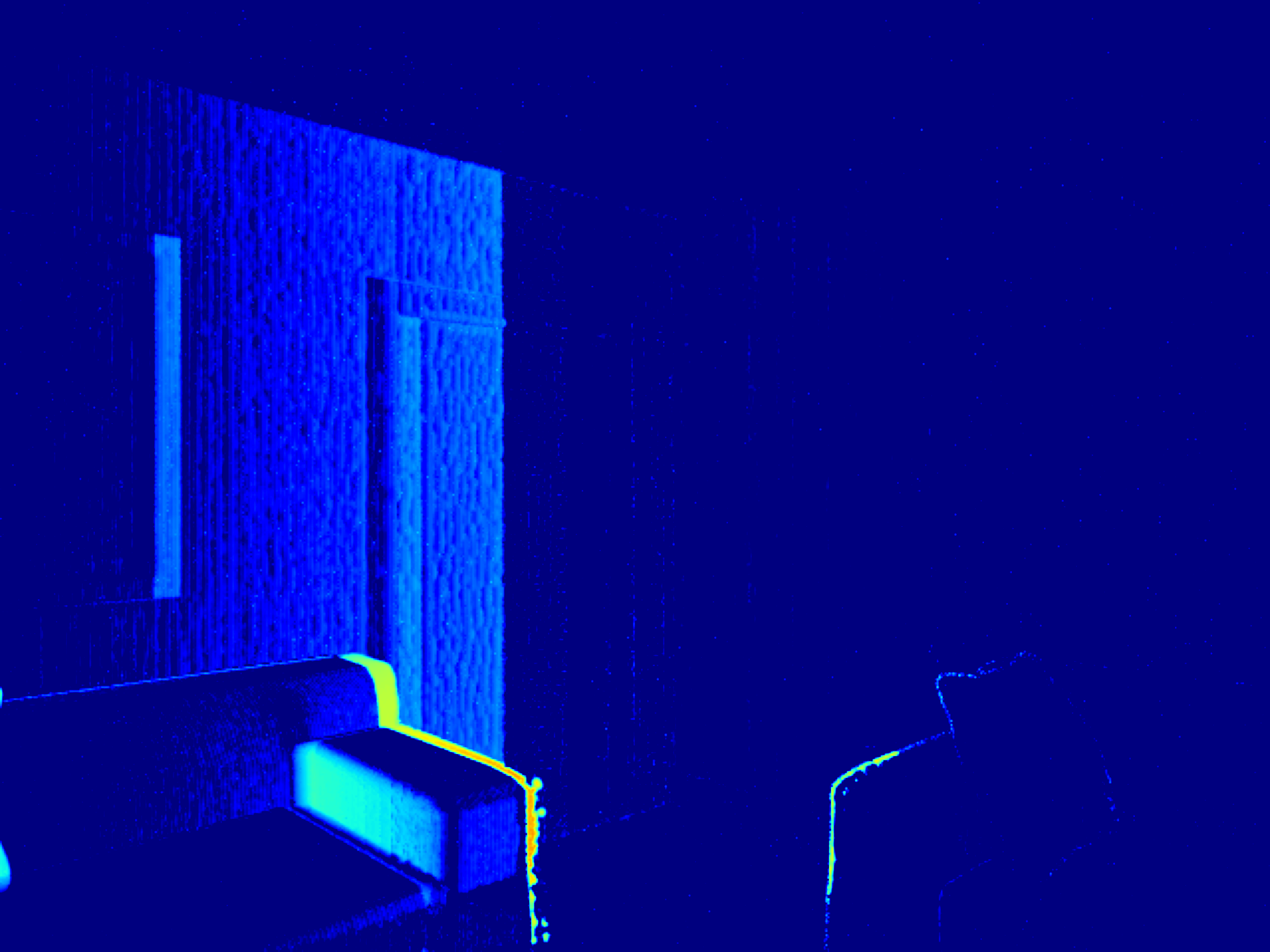}%
		\end{minipage}
	}%
    %\hspace{-4mm}
    \subfloat{%
		\begin{minipage}[b]{0.25\linewidth}
			\centering
			\includegraphics[width=1\linewidth]{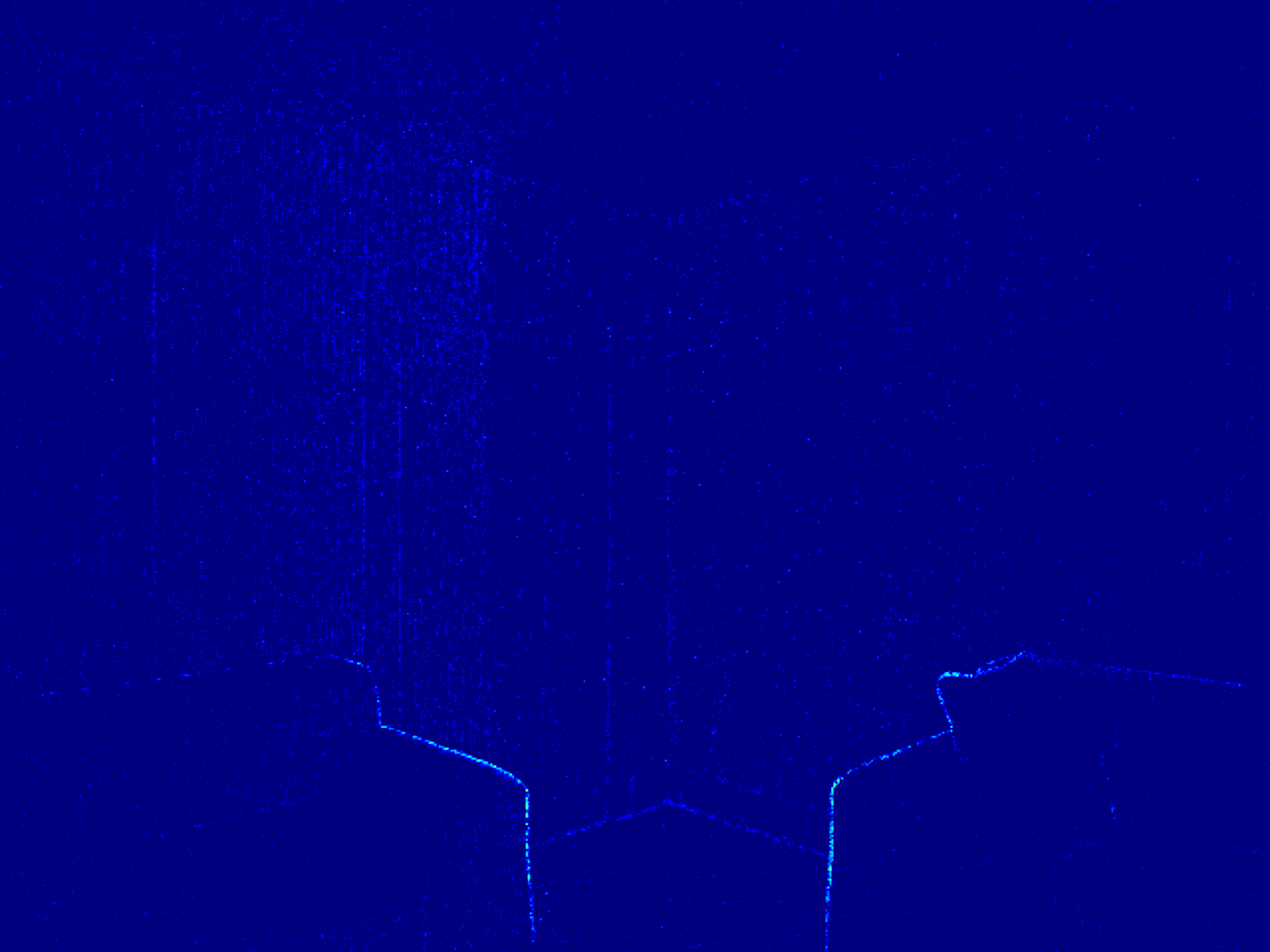}%
		\end{minipage}
	}\\
	\vspace{-2mm}%
    % 第3行
    \subfloat{%
        \hspace{-5mm}%
        \rotatebox{90}{\scriptsize{~~~~~~~~~~~~~~\textbf{office\_0}}}
		\begin{minipage}[b]{0.25\linewidth}
			\centering
			\includegraphics[width=1\linewidth]{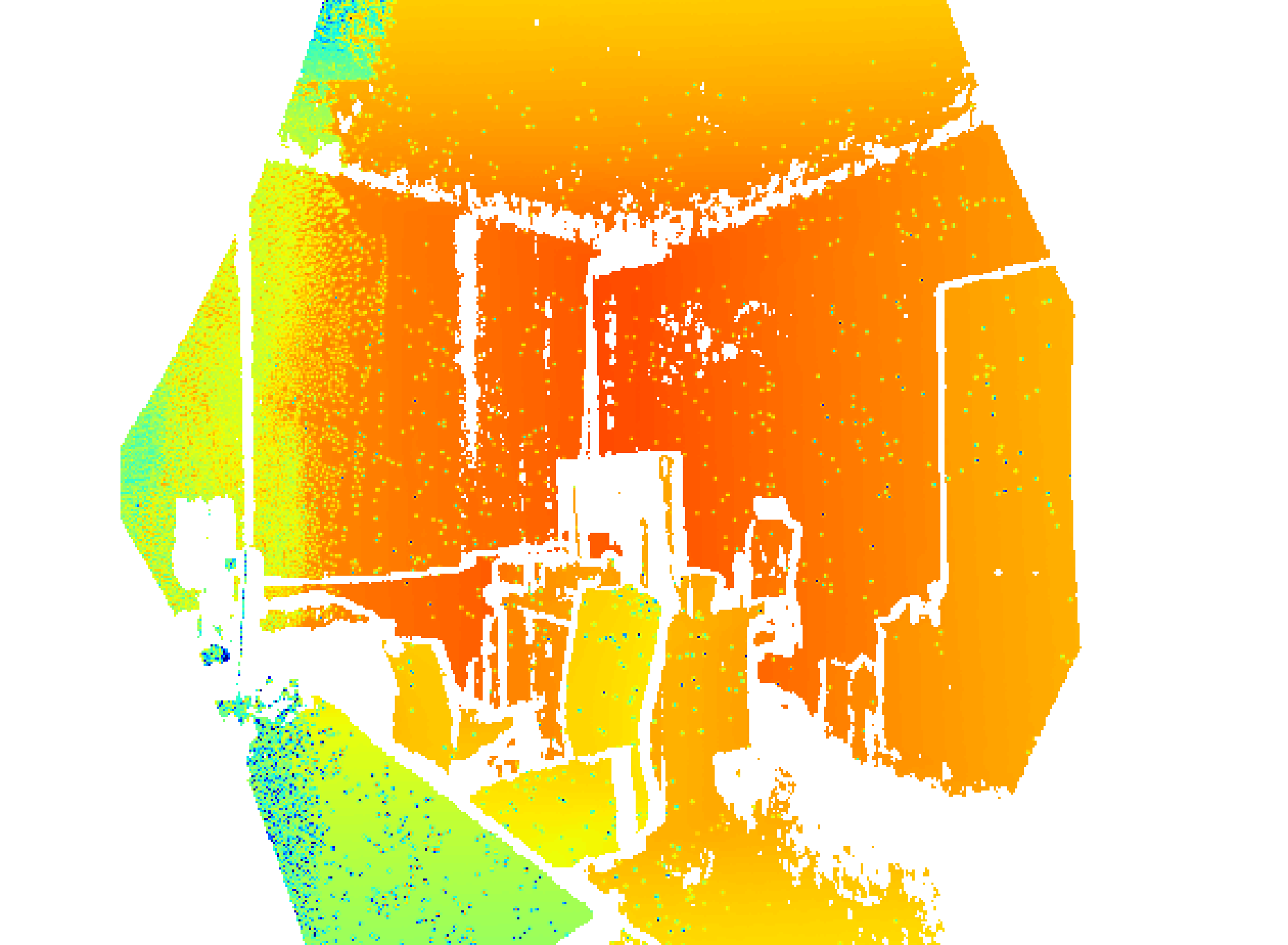}%
		\end{minipage}
	}%
    %\hspace{-4mm}
	\subfloat{%
		\begin{minipage}[b]{0.25\linewidth}
			\centering
			\includegraphics[width=1\linewidth]{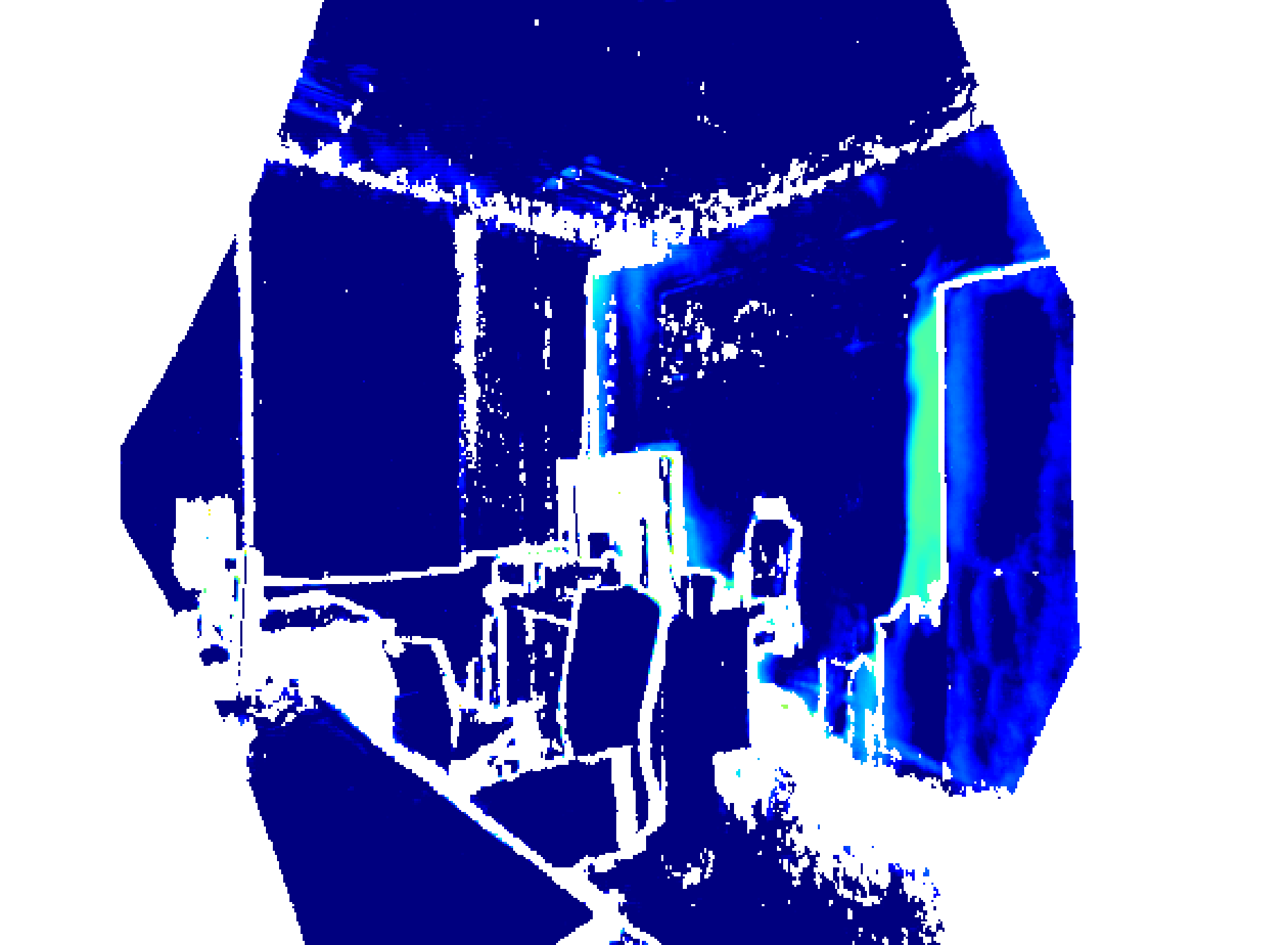}%
		\end{minipage}
	}%
    %\hspace{-4mm}
	\subfloat{%
		\begin{minipage}[b]{0.25\linewidth}
			\centering
			\includegraphics[width=1\linewidth]{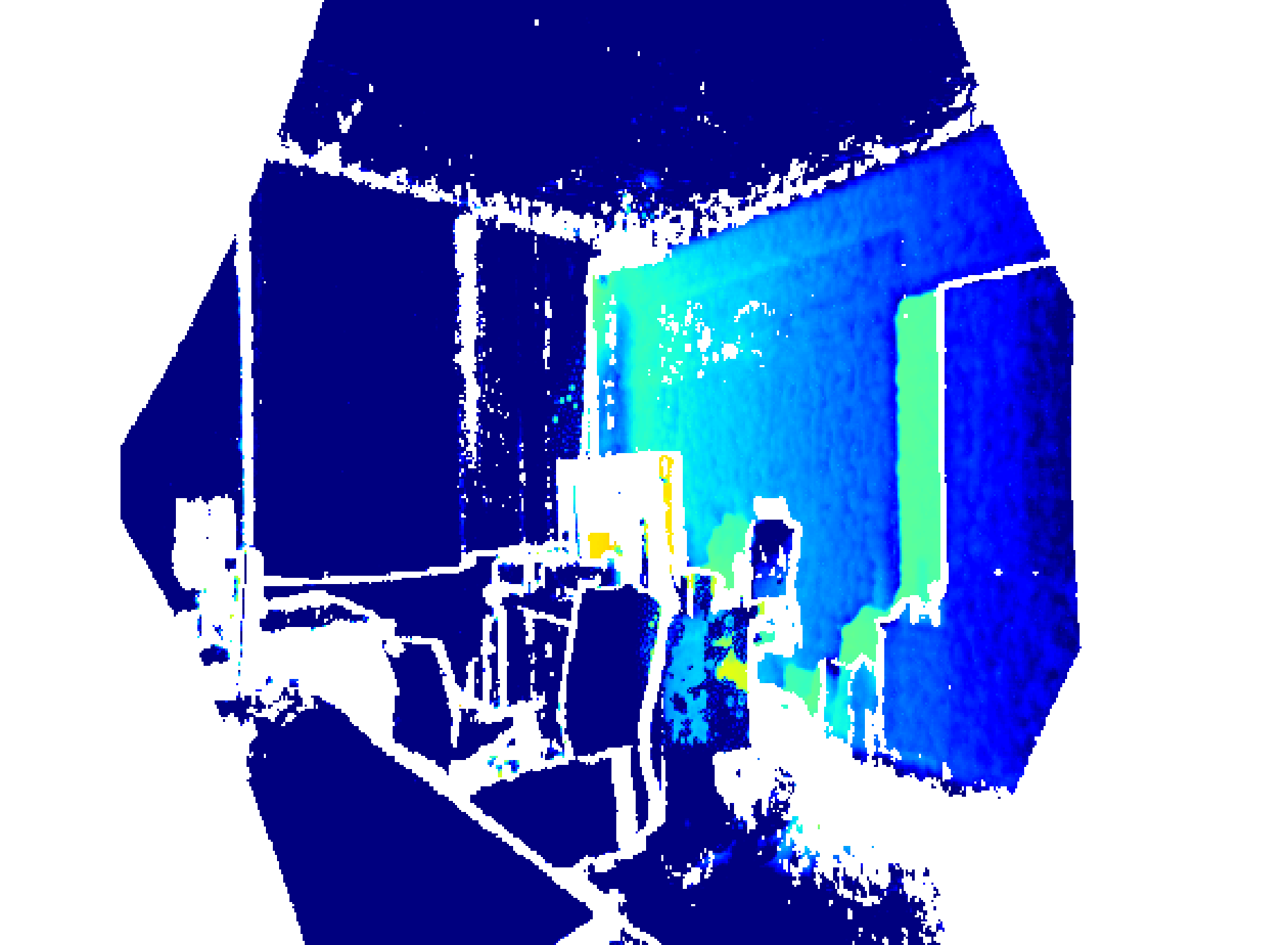}%
		\end{minipage}
	}%
    %\hspace{-4mm}
    \subfloat{%
		\begin{minipage}[b]{0.25\linewidth}
			\centering
			\includegraphics[width=1\linewidth]{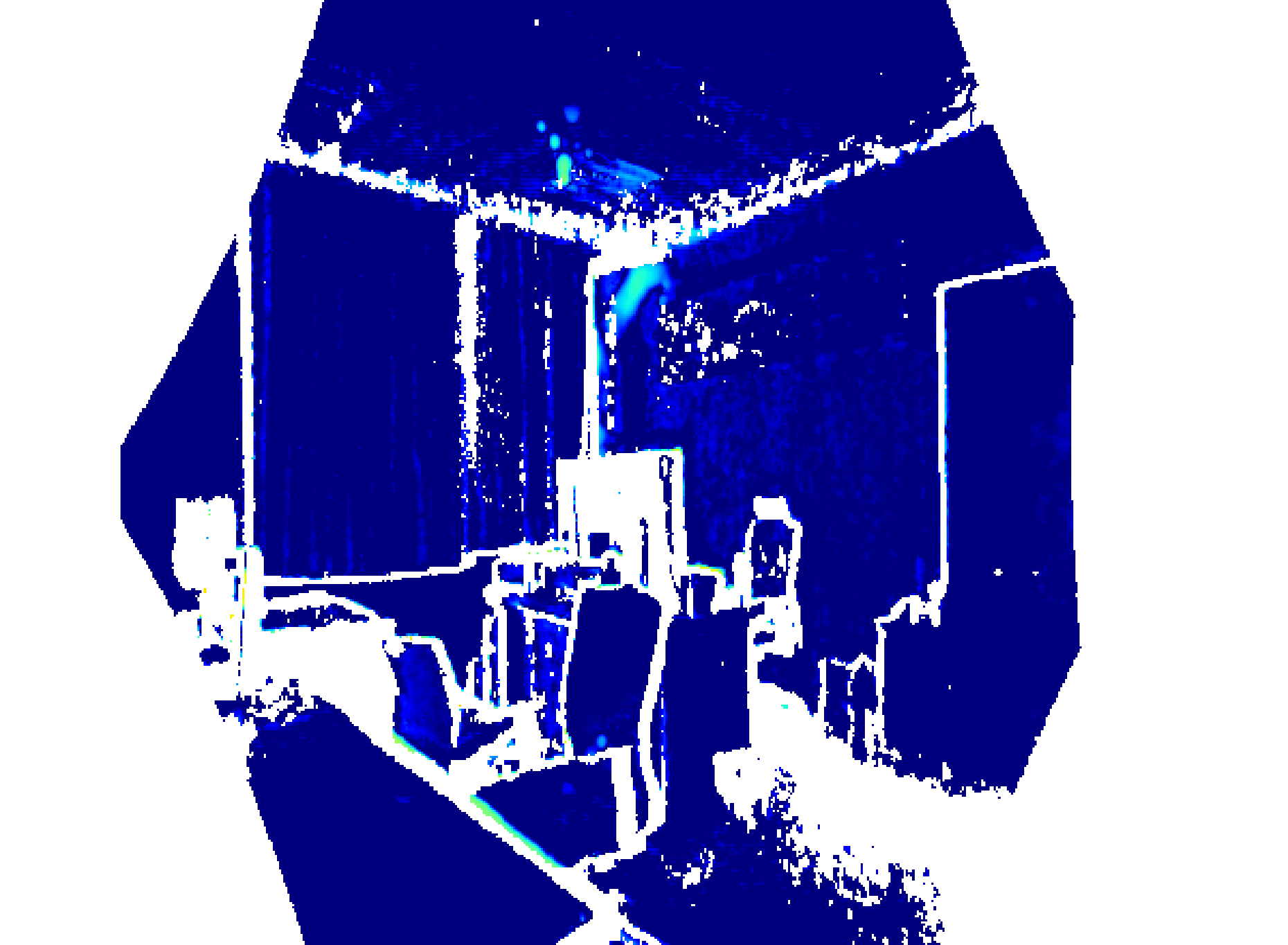}%
		\end{minipage}
	}\\
	\vspace{-2mm}
    % 第4行
    \subfloat{%
        \hspace{-5mm}%
        \rotatebox{90}{\scriptsize{~~~~~~~~~~~~~~\textbf{office\_1}}}
		\begin{minipage}[b]{0.25\linewidth}
			\centering
			\includegraphics[width=1\linewidth]{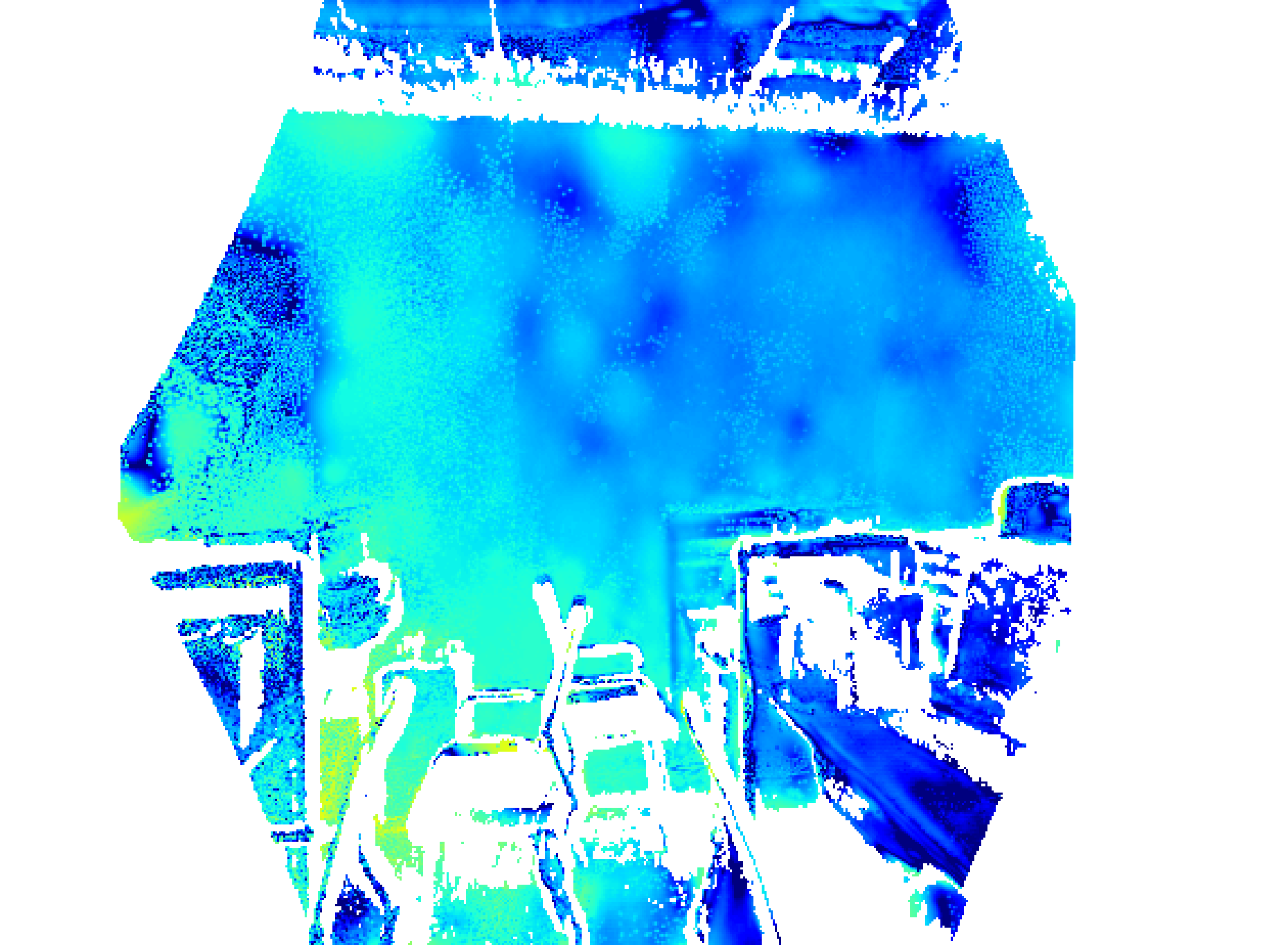}%
		\end{minipage}
	}%
    %\hspace{-4mm}
	\subfloat{%
		\begin{minipage}[b]{0.25\linewidth}
			\centering
			\includegraphics[width=1\linewidth]{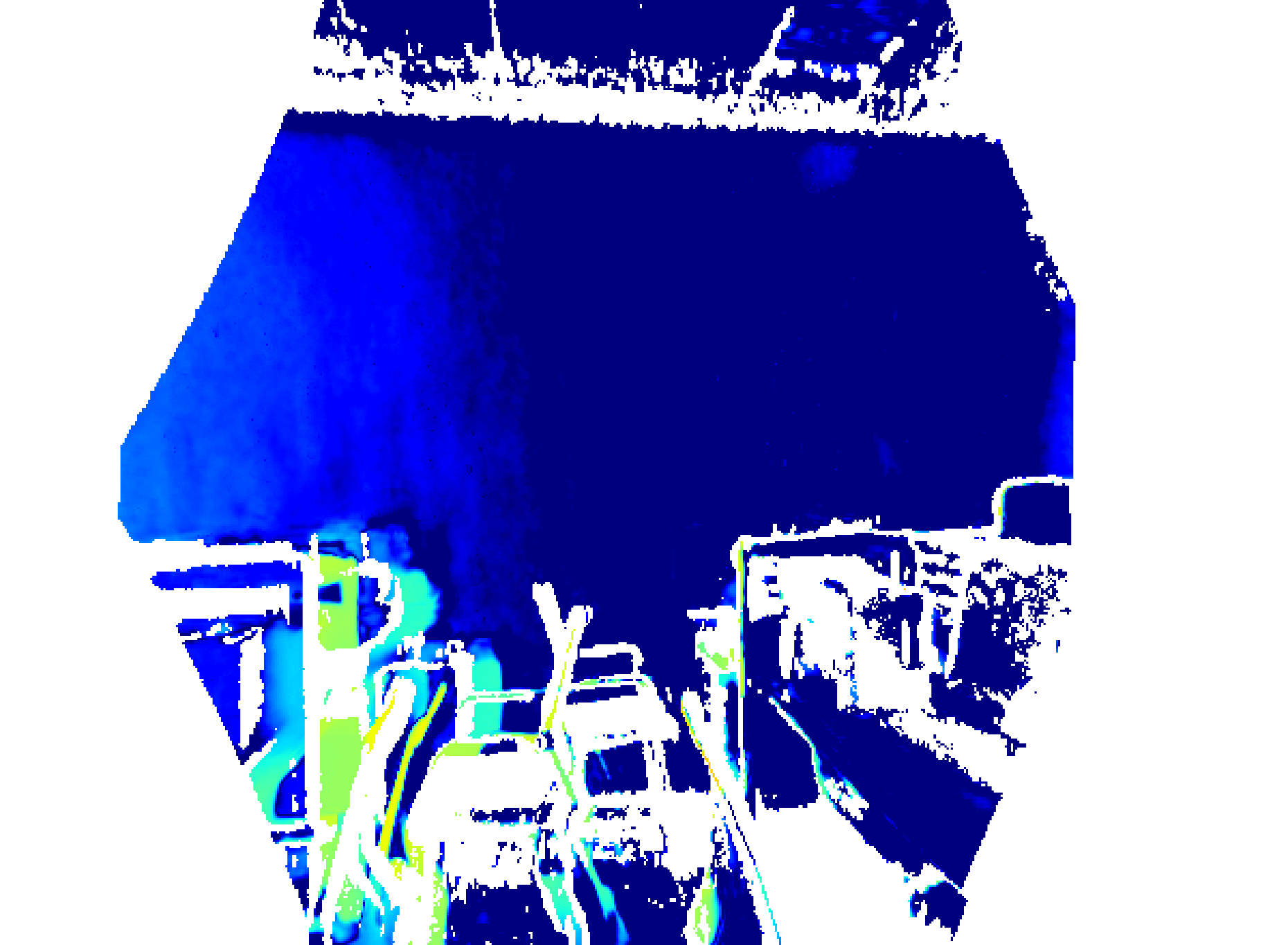}%
		\end{minipage}
	}%
    %\hspace{-4mm}
	\subfloat{%
		\begin{minipage}[b]{0.25\linewidth}
			\centering
			\includegraphics[width=1\linewidth]{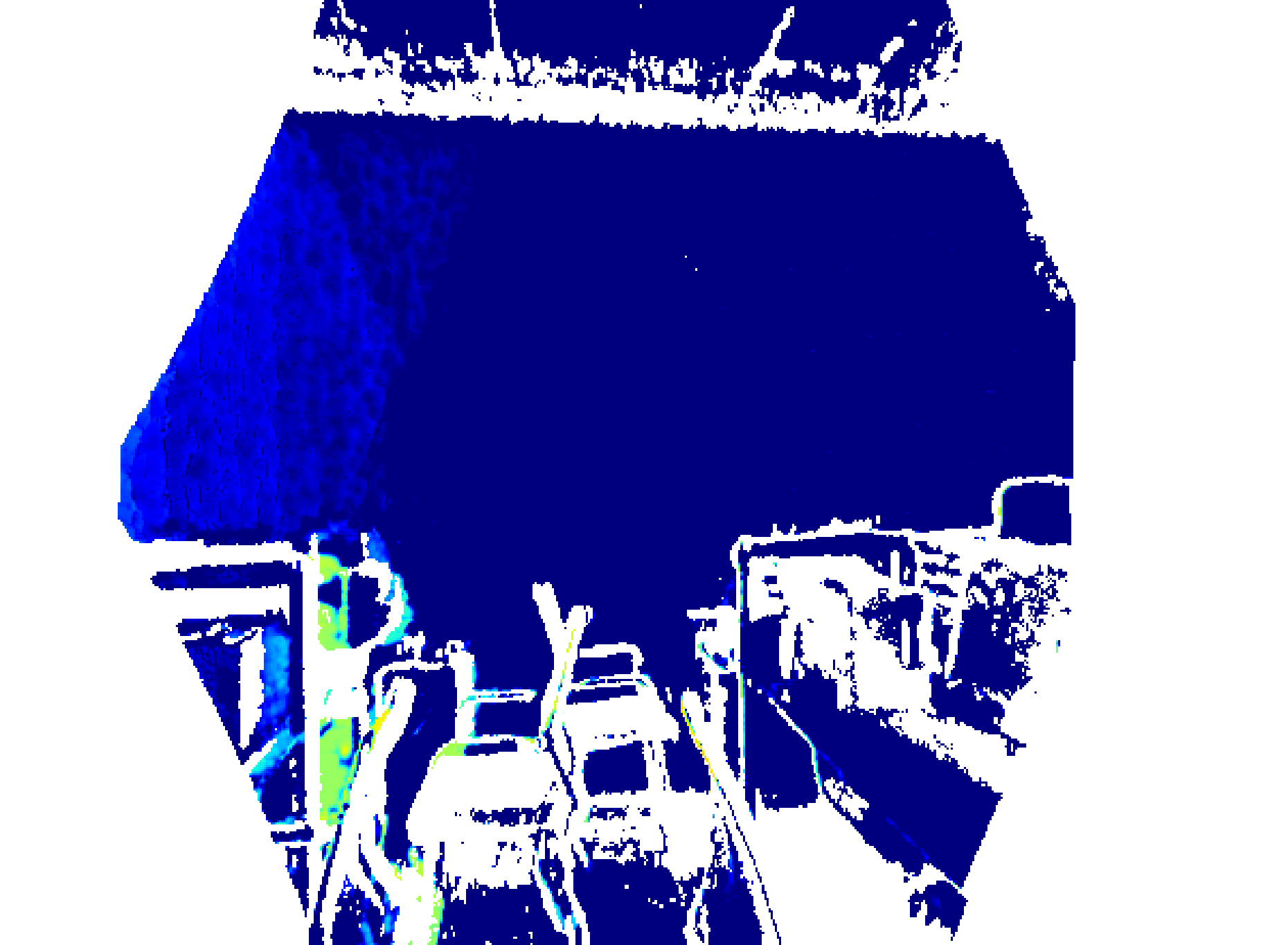}%
		\end{minipage}
	}%
    %\hspace{-4mm}
    \subfloat{%
		\begin{minipage}[b]{0.25\linewidth}
			\centering
			\includegraphics[width=1\linewidth]{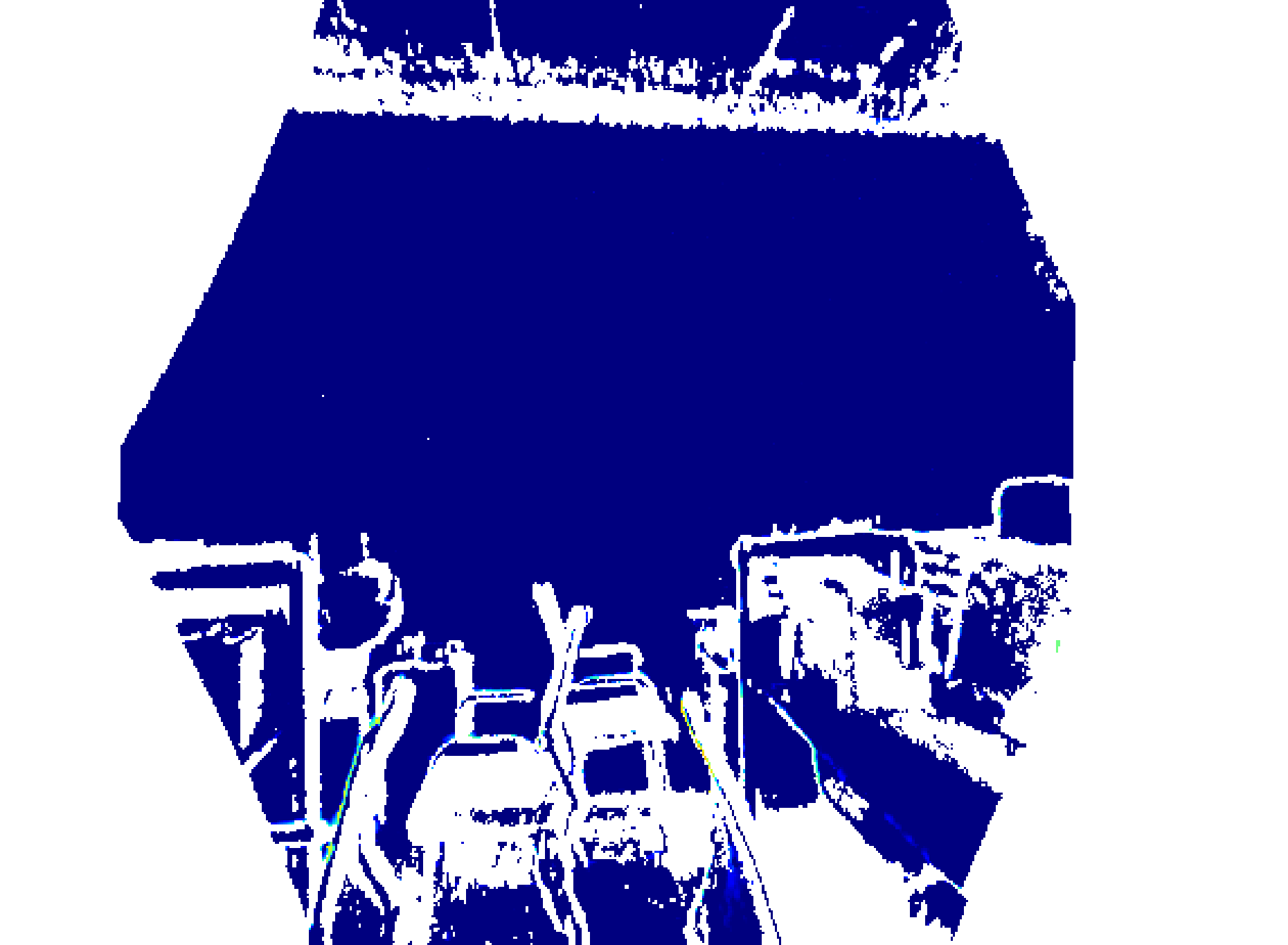}%
		\end{minipage}
	}\\
	\vspace{-2mm}
	%\setcounter{subfloat}{0}
    % 第5行
    \subfloat{%
        \hspace{-5mm}%
		\rotatebox{90}{\scriptsize{~~~~~~~~~~~~~~\textbf{hotel\_0}}}
		\begin{minipage}[t]{0.25\linewidth}
			\centering
			\includegraphics[width=1\linewidth]{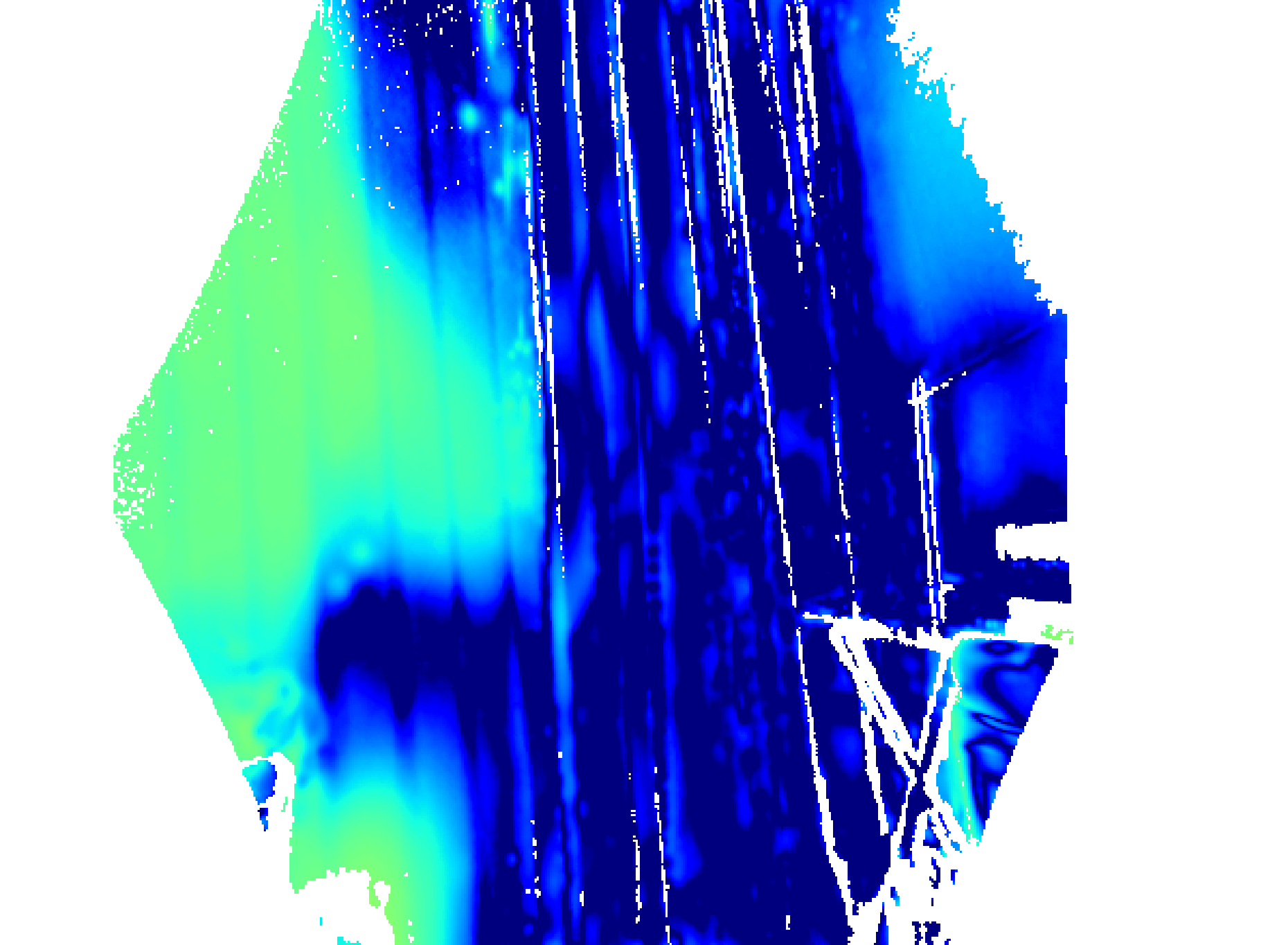}%
		\end{minipage}
	}%
	\subfloat{%
		\begin{minipage}[t]{0.25\linewidth}
			\centering
			\includegraphics[width=1\linewidth]{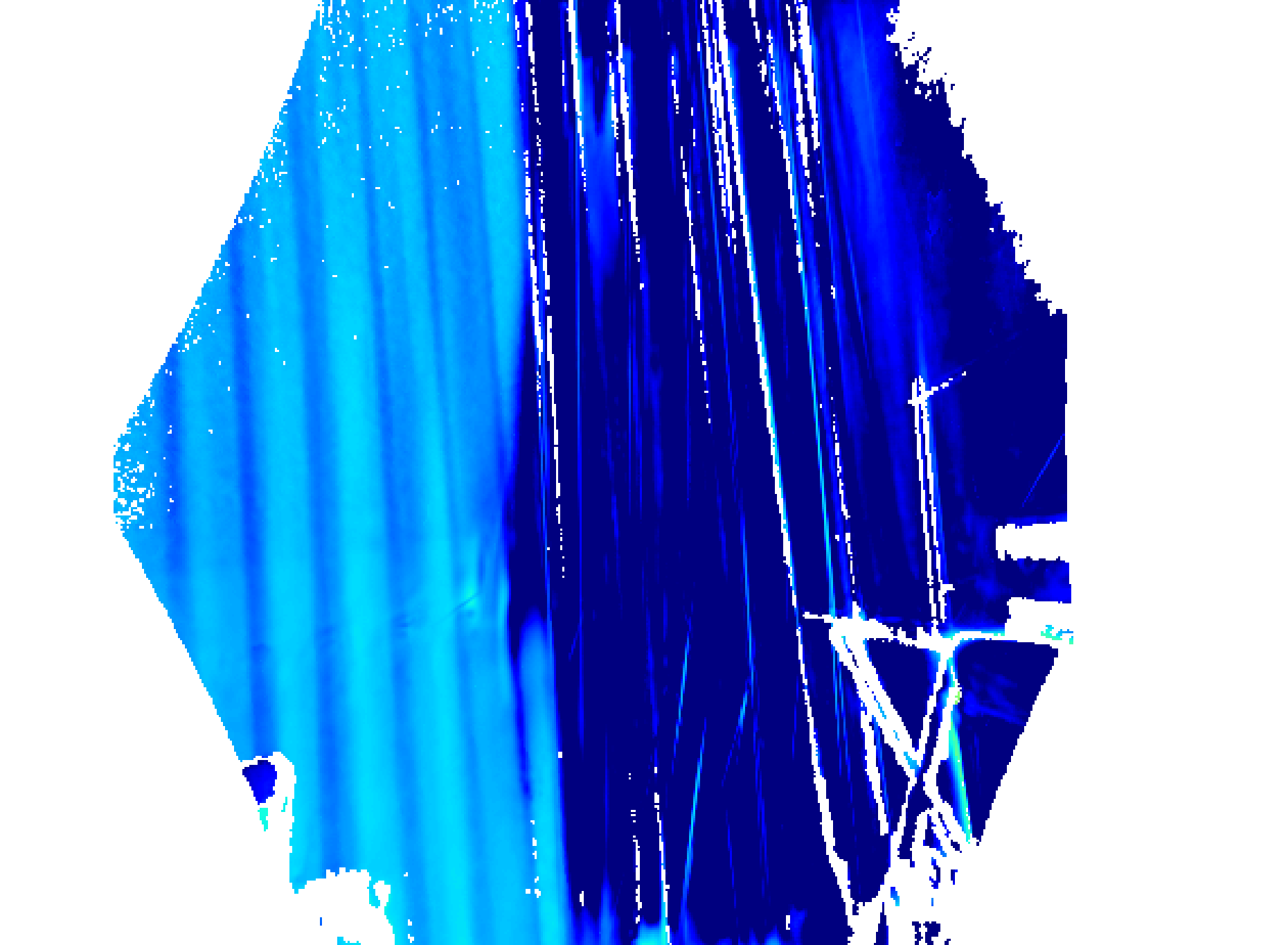}%
		\end{minipage}
	}%
	\subfloat{%
		\begin{minipage}[t]{0.25\linewidth}
			\centering
			\includegraphics[width=1\linewidth]{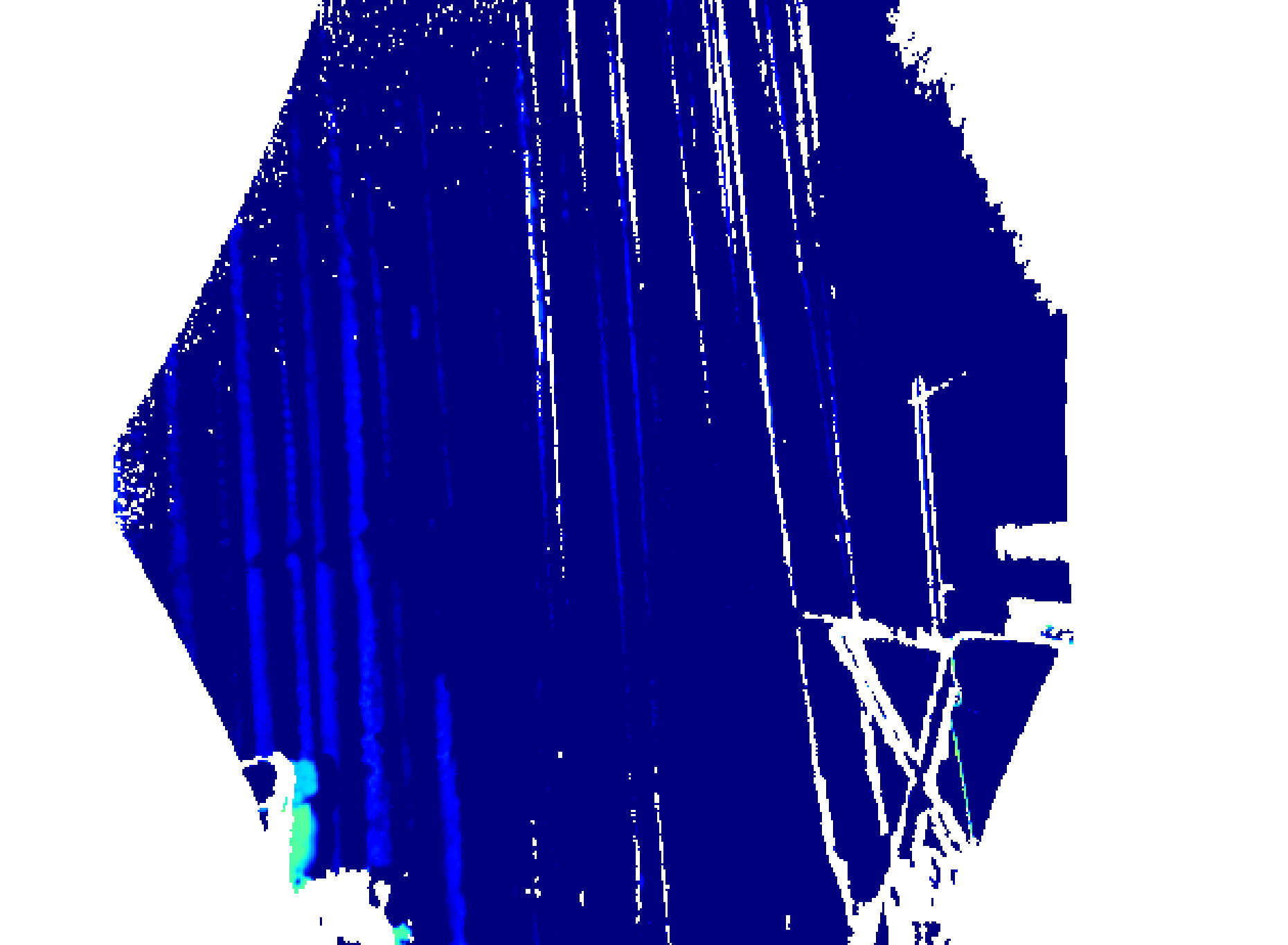}%
		\end{minipage}
	}%
    \subfloat{%
		\begin{minipage}[t]{0.25\linewidth}
			\centering
			\includegraphics[width=1\linewidth]{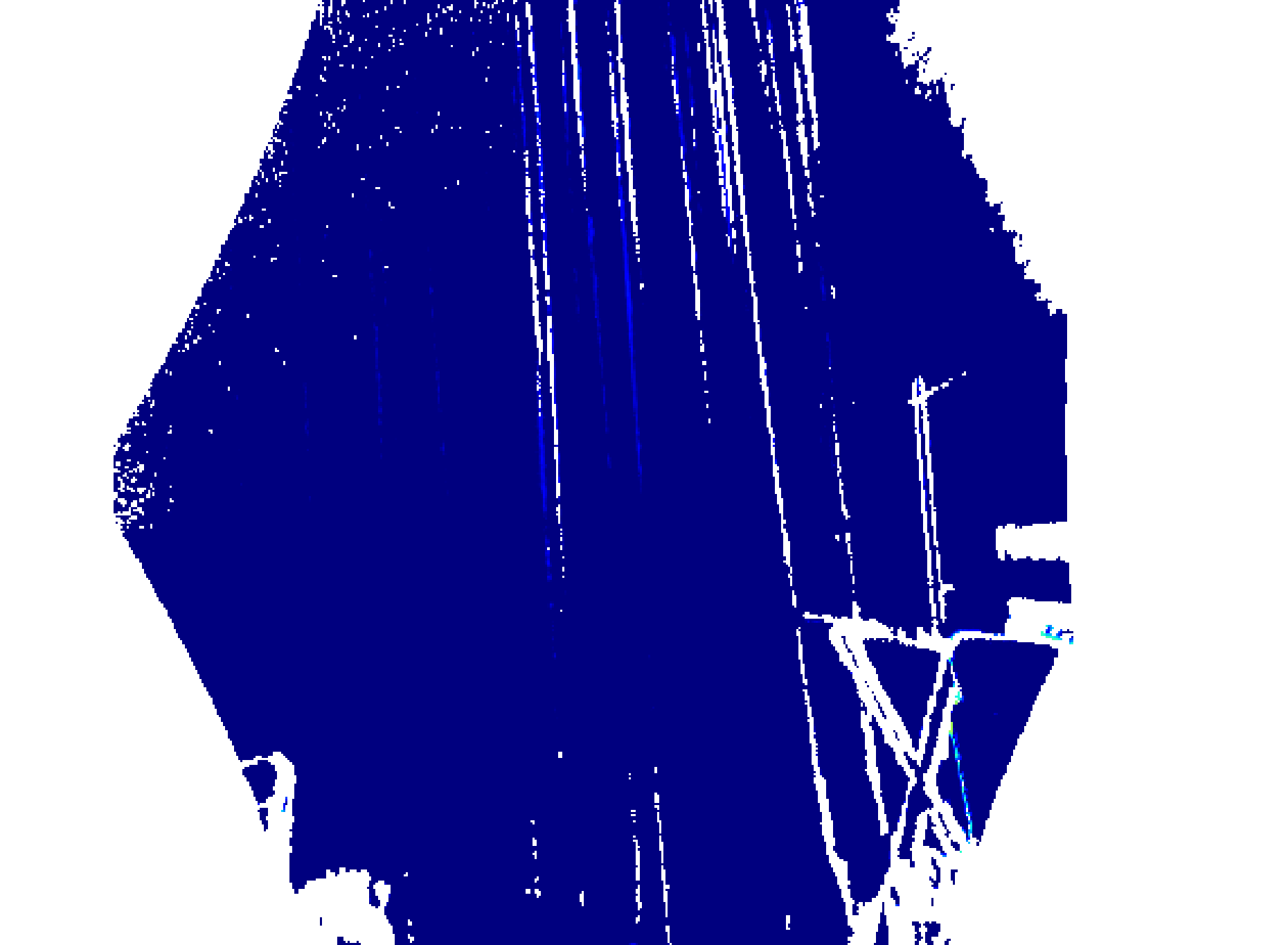}%
		\end{minipage}
	}\\
	\vspace{-2mm}
    %第6行
    \subfloat[MonoGS]{%
        \hspace{-5mm}%
		\rotatebox{90}{\scriptsize{~~~~~~~~~~~~~~\textbf{labor\_0}}}
		\begin{minipage}[t]{0.25\linewidth}
			\centering
			\includegraphics[width=1\linewidth]{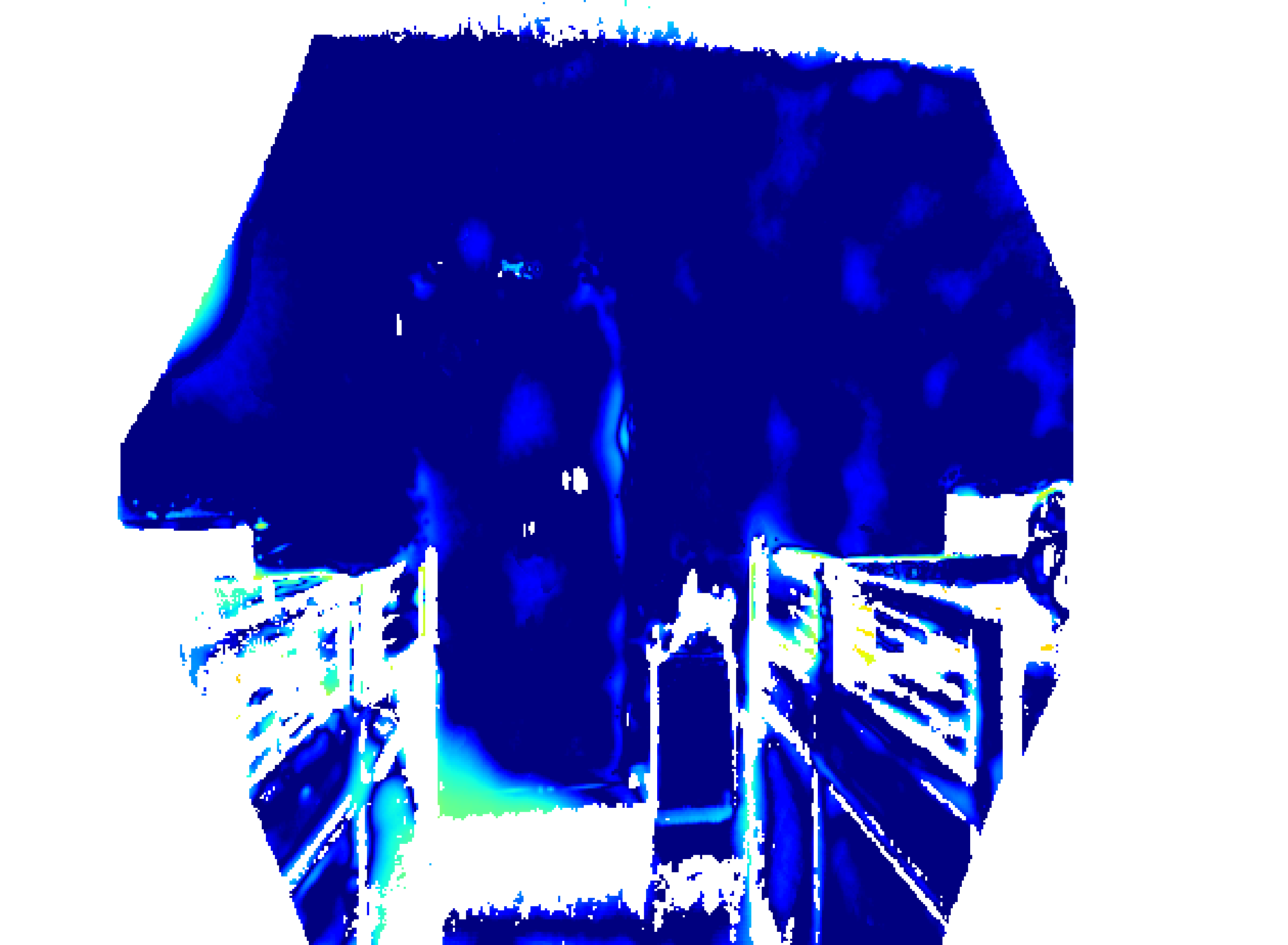}%
		\end{minipage}
	}%
	\subfloat[Gaussian-SLAM]{%
		\begin{minipage}[t]{0.25\linewidth}
			\centering
			\includegraphics[width=1\linewidth]{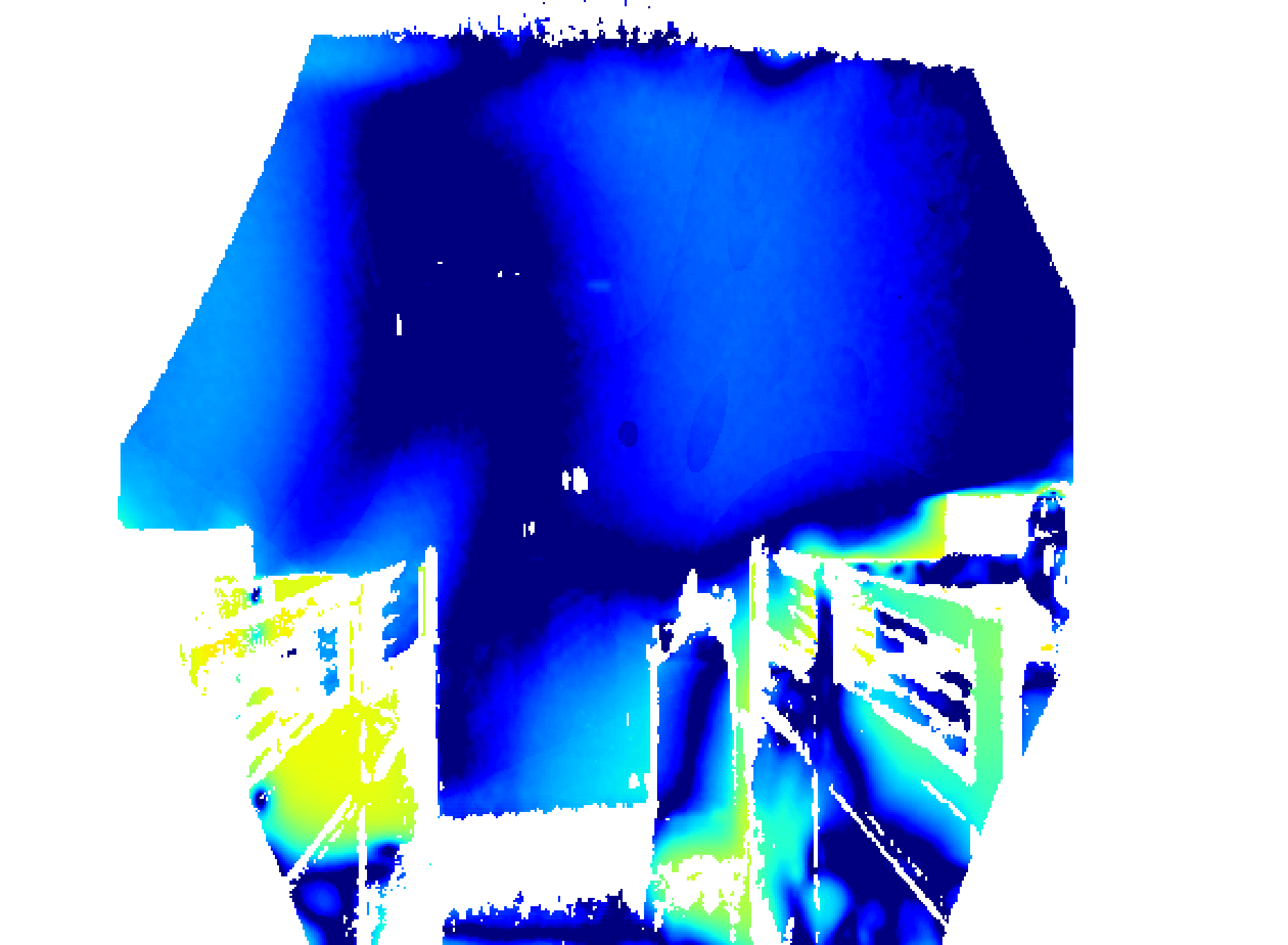}%
		\end{minipage}
	}%
	\subfloat[SplaTAM]{%
		\begin{minipage}[t]{0.25\linewidth}
			\centering
			\includegraphics[width=1\linewidth]{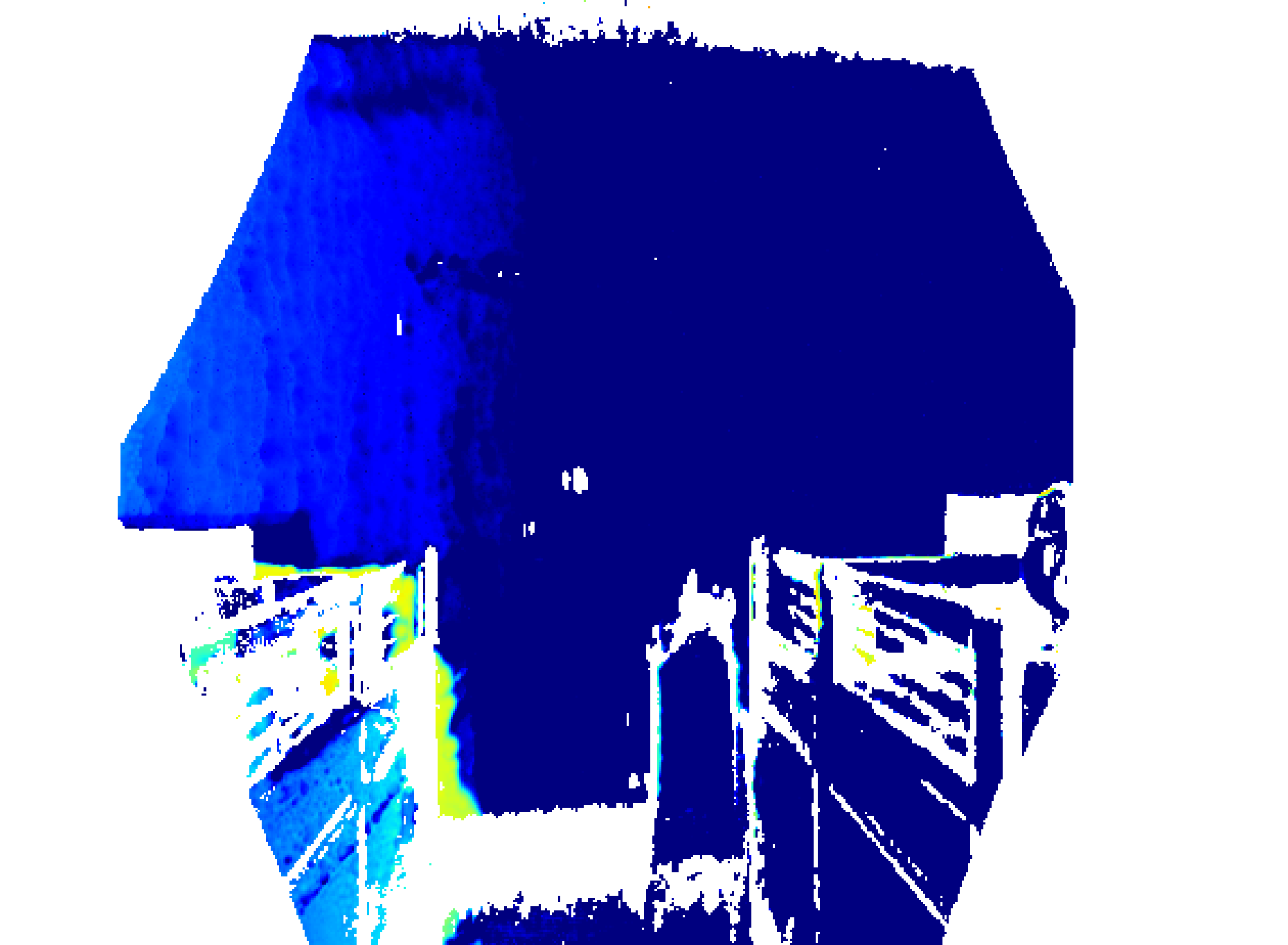}%
		\end{minipage}
	}%
    \subfloat[Ours]{%
		\begin{minipage}[t]{0.25\linewidth}
			\centering
			\includegraphics[width=1\linewidth]{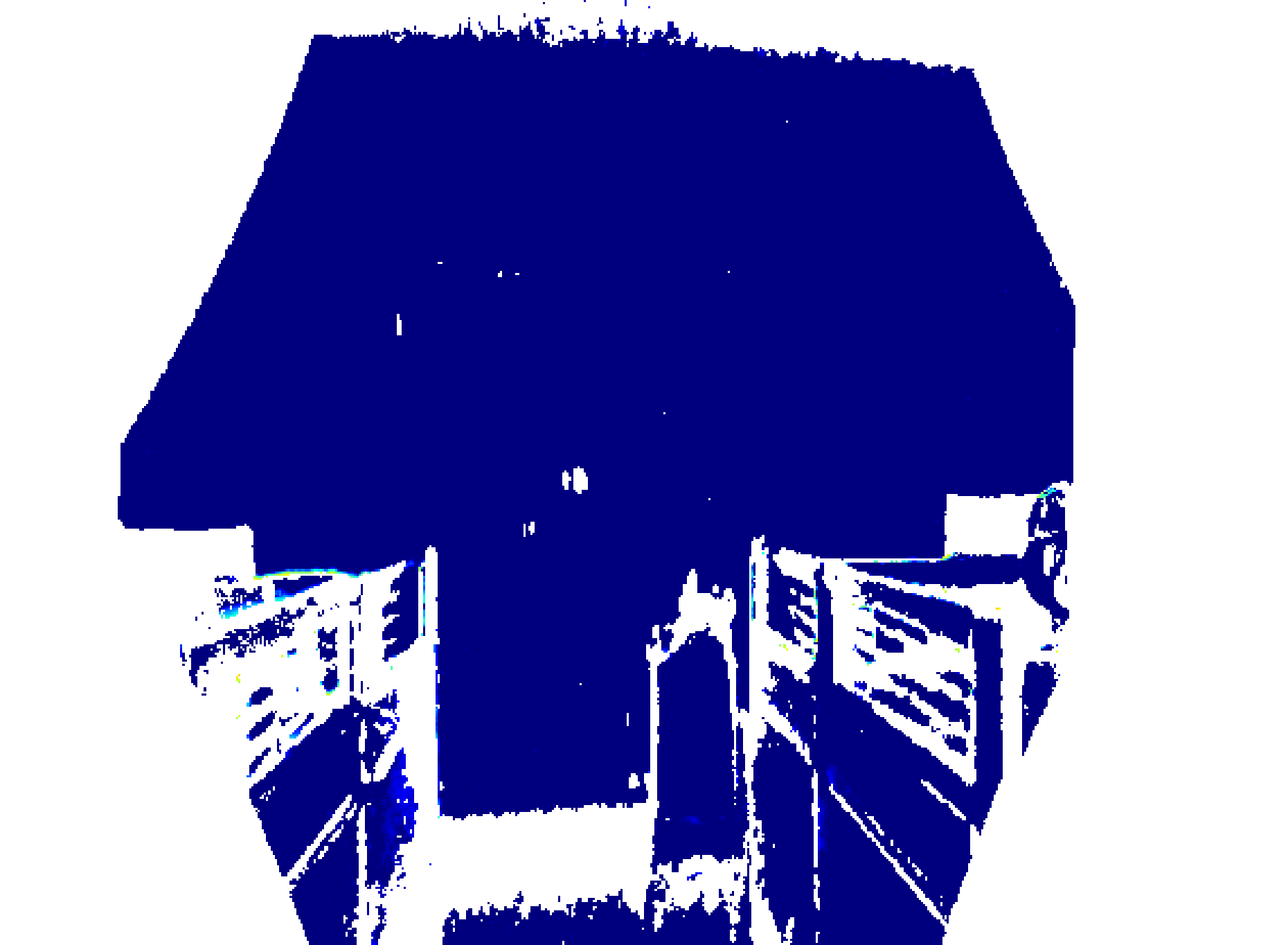}%
		\end{minipage}
	}
	\caption{\textbf{Comparison of depth error maps on virtual and real-world datasets.} Depth error maps obtained by calculating the differences between the rendered images and the ground truth are attached for better comparison. In these maps, shades of blue or cooler tones indicate smaller differences, while reds or warmer tones signify larger discrepancies.}
	\label{diff depth}%
\end{figure*}

% 真实数据集PSNR、SSIM、LPIPS、Depth L1指标对比
\begin{table*}[ht]
    \centering
    \captionsetup[table*]{singlelinecheck=off}
	\renewcommand{\arraystretch}{1.2} %rows, default value is 1.0
	\setlength{\tabcolsep}{4pt}
    \resizebox{\textwidth}{!}{
        \begin{tabular}{c|c|ccccccccc}
        \toprule
        Method                  & Metric            & \multicolumn{1}{l}{office\_0} & \multicolumn{1}{l}{office\_1} & \multicolumn{1}{l}{office\_2} & \multicolumn{1}{l}{hotel\_0} & \multicolumn{1}{l}{hotel\_1} & \multicolumn{1}{l}{labor\_0} & \multicolumn{1}{l}{labor\_1} & \multicolumn{1}{l}{labor\_2} & \multicolumn{1}{l}{Avg.} \\
        \hline
        \multirow{4}{*}{MonoGS\cite{Matsuki:Murai:etal:CVPR2024}} & 
        PSNR{[}dB{]} $\uparrow$  & 11.86  & 21.23  & 14.43  & 21.62   & 19.52   & 12.71  & 16.27 & 17.81 & 16.931 \\
        & SSIM$\uparrow$   & 0.092   & 0.74   & 0.337  & 0.799   & 0.717   & 0.568  & 0.610 & 0.704 & 0.571      \\
        & LPIPS$\downarrow$  & 0.96  & 0.439  & 0.850    & 0.345   & 0.503    & 0.637   & 0.568& 0.376& 0.585   \\
        & Depth L1{[}cm{]}$\downarrow$  & 69.6    & 12.02     & 33.07     & 6.79   & 12.33  & 48.90  & 23.95& 11.59& 27.281 \\
        \hline
        \multirow{4}{*}{Gaussian-SLAM\cite{yugay2023gaussianslam}} & 
        PSNR{[}dB{]} $\uparrow$  & 29.49  & 30.59  & 29.15  & 28.43   & 24.96   & 16.12  & 24.93 & 24.95& 26.078 \\
        & SSIM$\uparrow$   & 0.954   & 0.954   & 0.917 & 0.937   & 0.889   & 0.591  & 0.903 & 0.890  & 0.879 \\
        & LPIPS$\downarrow$  & 0.136  & 0.132  & 0.21    & 0.194   & 0.345    & 0.647   & 0.216& 0.254& 0.267 \\
        & Depth L1{[}cm{]}$\downarrow$  & \textbf{0.23}    & \textbf{0.24}      & 0.52     & 0.33   & 0.95  & 0.92  & \textbf{0.26}& \textbf{0.22}& 0.458 \\
        \hline
        \multirow{4}{*}{SplaTAM\cite{keetha2024splatam}} & 
        PSNR{[}dB{]} $\uparrow$  & 34.29  & 33.49  & 33.95  & 30.83   & 32.49   & 32.08  & 30.79 & 33.21& 32.641 \\
        & SSIM$\uparrow$   & 0.987   & 0.983   & 0.983 & 0.971   & 0.977   & 0.990  & 0.977  & 0.986  & 0.982   \\
        & LPIPS$\downarrow$  & 0.024  & 0.036  & 0.039    & 0.051   & 0.049    & 0.017   & 0.030& 0.030 & 0.035  \\
        & Depth L1{[}cm{]}$\downarrow$  & 0.55    & 0.49      & 0.35     & 0.32   & 0.31  & 0.61 & 0.59& 0.35& 0.446 \\
        \hline
        \multirow{4}{*}{Ours} & 
        PSNR{[}dB{]} $\uparrow$  & \textbf{35.65}  & \textbf{35.68}  & \textbf{36.54}  & \textbf{33.99}   & \textbf{35.66}   & \textbf{32.86}  & \textbf{31.76}  & \textbf{34.90}& \textbf{34.63}\\
        & SSIM$\uparrow$   & \textbf{0.993}   &\textbf{ 0.992}   & \textbf{0.991} & \textbf{0.986}   & \textbf{0.989}   & \textbf{0.992}  & \textbf{0.990} & \textbf{0.992}& \textbf{0.991} \\
        & LPIPS$\downarrow$  & \textbf{0.016}  &\textbf{ 0.022}  & \textbf{0.018}    & \textbf{0.032}   & \textbf{0.032}    & \textbf{0.014}   & \textbf{0.018}  & \textbf{0.021}& \textbf{0.022} \\
        & Depth L1{[}cm{]}$\downarrow$  & 0.29    & 0.28   & \textbf{0.24}     & \textbf{0.19}   & \textbf{0.21}  & \textbf{0.50}  & 0.36 & 0.28& \textbf{0.294}\\
        \bottomrule
        \end{tabular}}
\caption{\textbf{Rendering and reconstruction performance in real-world sequences.} Results with best accuracy are highlighted by \textbf{bold} font.}
\label{real data}
\end{table*}

% 真实数据集渲染结果对比
\begin{figure*}[htbp]
	\centering
    \captionsetup[subfloat]{labelformat=empty}
	\subfloat{%
        \hspace{-5mm}%
        \rotatebox{90}{\scriptsize{~~~~~~~~~~\textbf{office\_0}}}
		\begin{minipage}[b]{0.20\linewidth}
			\includegraphics[width=1\linewidth]{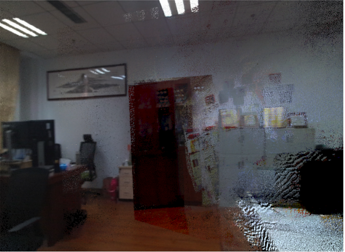}%
		\end{minipage}
	}
	\subfloat{%
		\begin{minipage}[b]{0.20\linewidth}
			\includegraphics[width=1\linewidth]{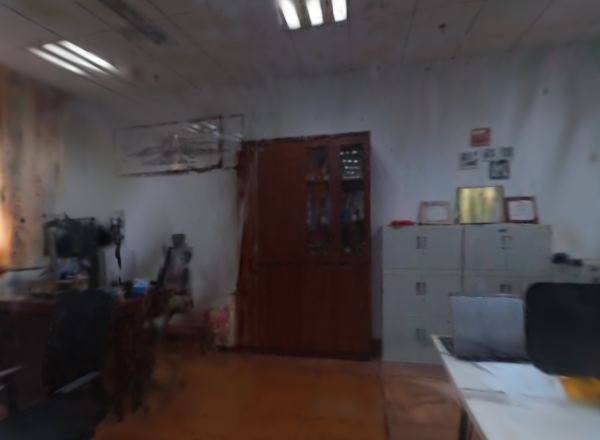}%
		\end{minipage}
	}
	\subfloat{%
		\begin{minipage}[b]{0.20\linewidth}
			\includegraphics[width=1\linewidth]{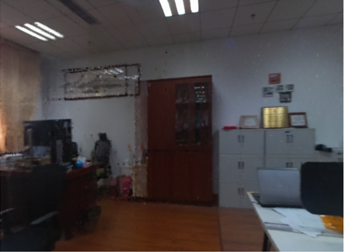}%
		\end{minipage}
	}
    \subfloat{%
		\begin{minipage}[b]{0.20\linewidth}
			\includegraphics[width=1\linewidth]{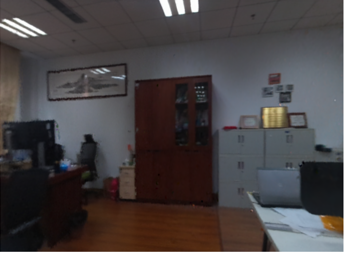}%
		\end{minipage}
	}
    \subfloat{%
		\begin{minipage}[b]{0.20\linewidth}
			\includegraphics[width=1\linewidth]{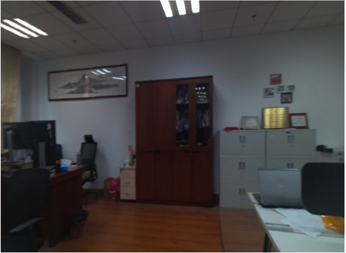}%
		\end{minipage}
	}\\
     % 第二行
	\vspace{-2mm}
    \subfloat{%
        \hspace{-5mm}%
        \rotatebox{90}{\scriptsize{~~~~~~~~~~\textbf{office\_1}}}
		\begin{minipage}[b]{0.20\linewidth}
			\includegraphics[width=1\linewidth]{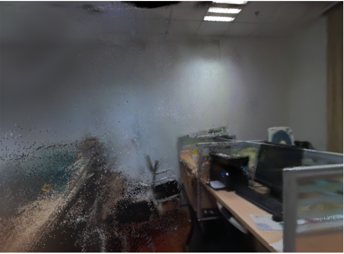}%
		\end{minipage}
	}
	\subfloat{%
		\begin{minipage}[b]{0.20\linewidth}
			\includegraphics[width=1\linewidth]{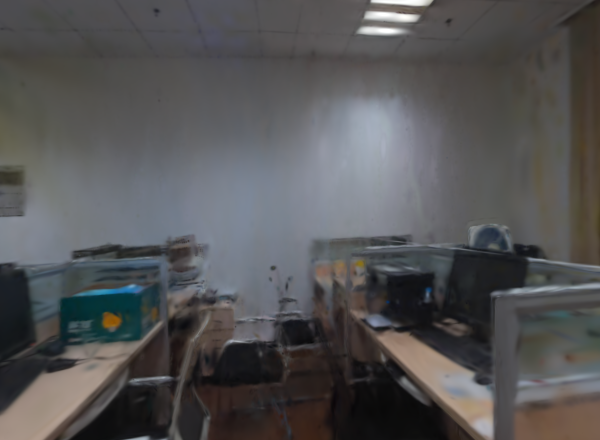}%
		\end{minipage}
	}
	\subfloat{%
		\begin{minipage}[b]{0.20\linewidth}
			\includegraphics[width=1\linewidth]{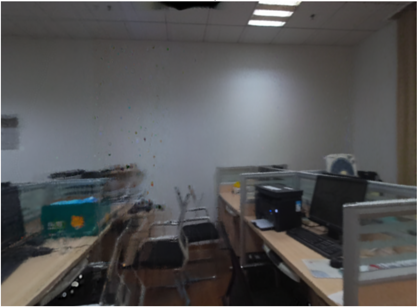}%
		\end{minipage}
	}
    \subfloat{%
		\begin{minipage}[b]{0.20\linewidth}
			\includegraphics[width=1\linewidth]{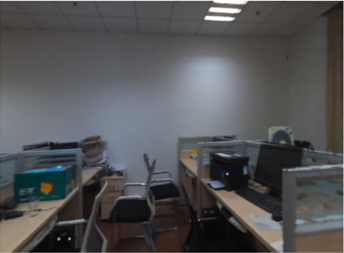}%
		\end{minipage}
	}
    \subfloat{%
		\begin{minipage}[b]{0.20\linewidth}
			\includegraphics[width=1\linewidth]{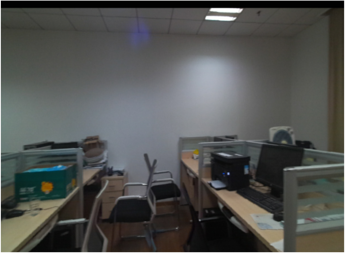}%
		\end{minipage}
	}\\
	\vspace{-2mm}
    % 第三行
    \subfloat{%
        \hspace{-5mm}%
        \rotatebox{90}{\scriptsize{~~~~~~~~~~\textbf{office\_2}}}
		\begin{minipage}[b]{0.20\linewidth}
			\includegraphics[width=1\linewidth]{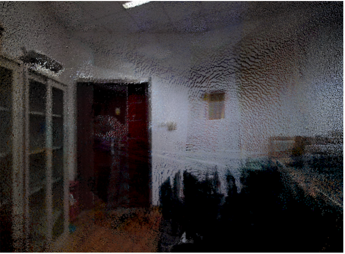}%
		\end{minipage}
	}
	\subfloat{%
		\begin{minipage}[b]{0.20\linewidth}
			\includegraphics[width=1\linewidth]{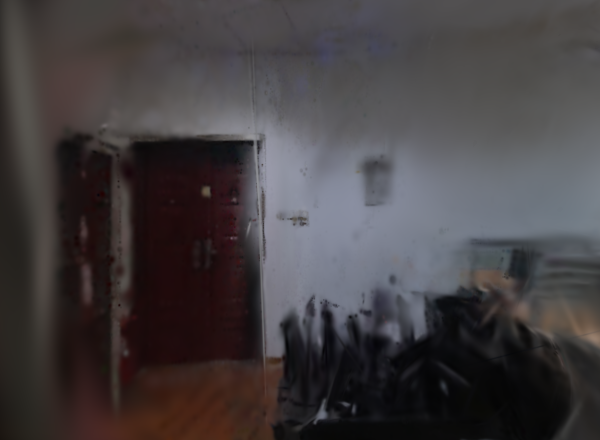}%
		\end{minipage}
	}
	\subfloat{%
		\begin{minipage}[b]{0.20\linewidth}
			\includegraphics[width=1\linewidth]{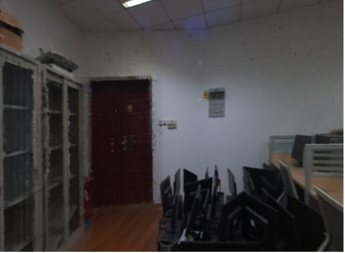}%
		\end{minipage}
	}
    \subfloat{%
		\begin{minipage}[b]{0.20\linewidth}
			\includegraphics[width=1\linewidth]{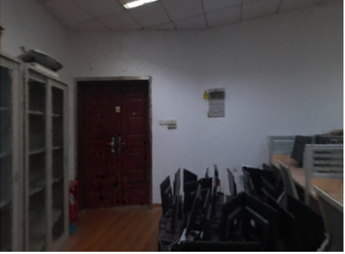}%
		\end{minipage}
	}
    \subfloat{%
		\begin{minipage}[b]{0.20\linewidth}
			\includegraphics[width=1\linewidth]{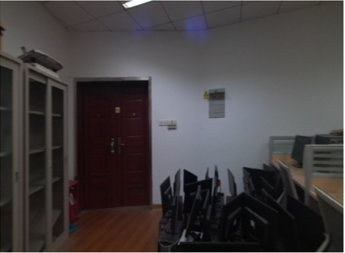}%
		\end{minipage}
	}\\
    \vspace{-2mm}
    %第四行
    \subfloat{%
        \hspace{-5mm}%
        \rotatebox{90}{\scriptsize{~~~~~~~~~~\textbf{labor\_0}}}
		\begin{minipage}[b]{0.20\linewidth}
			\includegraphics[width=1\linewidth]{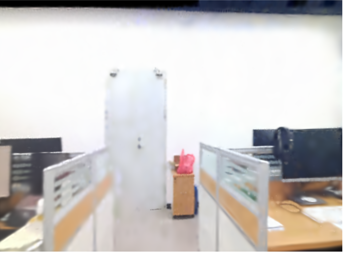}%
		\end{minipage}
	}
	\subfloat{%
		\begin{minipage}[b]{0.20\linewidth}
			\includegraphics[width=1\linewidth]{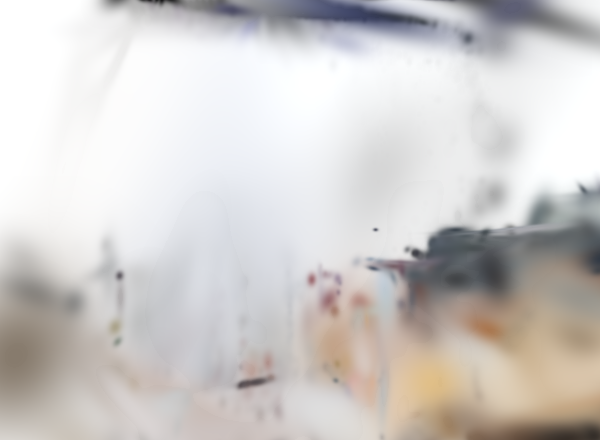}%
		\end{minipage}
	}
	\subfloat{%
		\begin{minipage}[b]{0.20\linewidth}
			\includegraphics[width=1\linewidth]{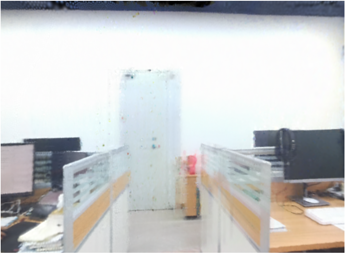}%
		\end{minipage}
	}
    \subfloat{%
		\begin{minipage}[b]{0.20\linewidth}
			\includegraphics[width=1\linewidth]{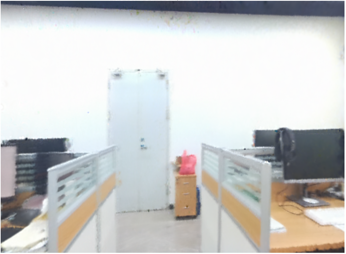}%
		\end{minipage}
	}
    \subfloat{%
		\begin{minipage}[b]{0.20\linewidth}
			\includegraphics[width=1\linewidth]{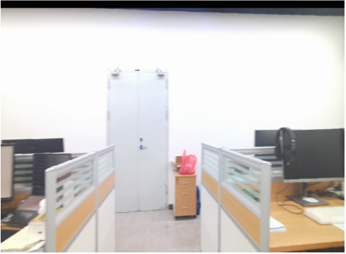}%
		\end{minipage}
	}\\
	\vspace{-2mm}
    %第五行
    \subfloat{%
        \hspace{-5mm}%
        \rotatebox{90}{\scriptsize{~~~~~~~~~~\textbf{labor\_1}}}
		\begin{minipage}[b]{0.20\linewidth}
			\includegraphics[width=1\linewidth]{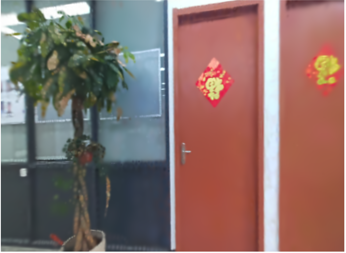}%
		\end{minipage}
	}
	\subfloat{%
		\begin{minipage}[b]{0.20\linewidth}
			\includegraphics[width=1\linewidth]{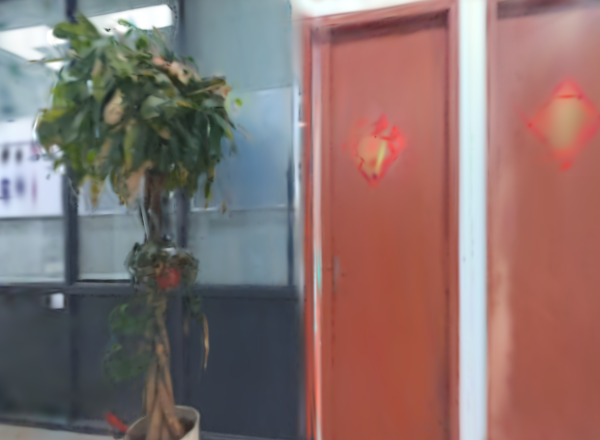}%
		\end{minipage}
	}
	\subfloat{%
		\begin{minipage}[b]{0.20\linewidth}
			\includegraphics[width=1\linewidth]{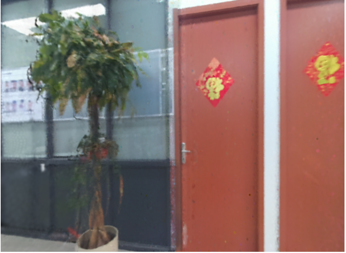}%
		\end{minipage}
	}
    \subfloat{%
		\begin{minipage}[b]{0.20\linewidth}
			\includegraphics[width=1\linewidth]{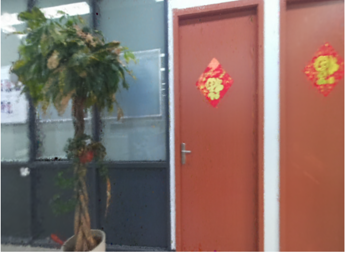}%
		\end{minipage}
	}
    \subfloat{%
		\begin{minipage}[b]{0.20\linewidth}
			\includegraphics[width=1\linewidth]{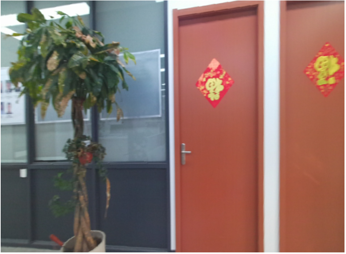}%
		\end{minipage}
	}\\
	\vspace{-2mm}
    % 最后一行
    \subfloat[MonoGS]{%
        \hspace{-5mm}%
		\rotatebox{90}{\scriptsize{~~~~~~~~~~\textbf{labor\_2}}}
		\begin{minipage}[b]{0.20\linewidth}
			\includegraphics[width=1\linewidth]{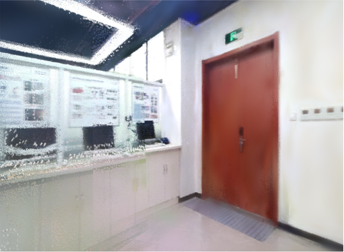}%
		\end{minipage}
	}
	\subfloat[Gaussian-SLAM]{%
		\begin{minipage}[b]{0.20\linewidth}
			\includegraphics[width=1\linewidth]{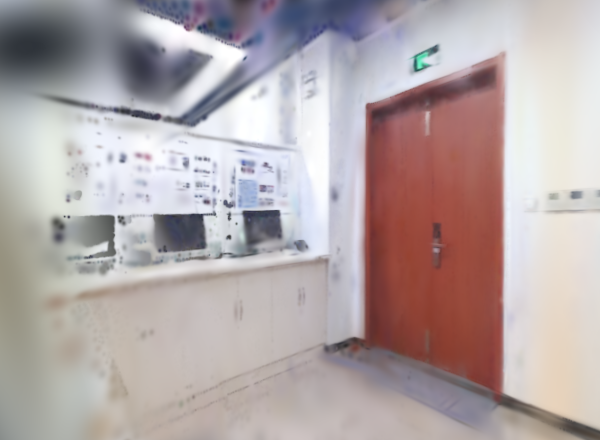}%
		\end{minipage}
	}
	\subfloat[SplaTAM]{%
		\begin{minipage}[b]{0.20\linewidth}
			\includegraphics[width=1\linewidth]{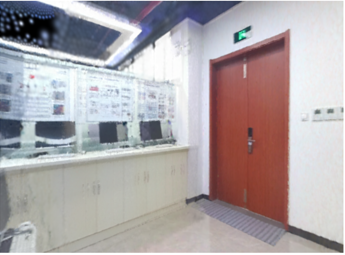}%
		\end{minipage}
	}
    \subfloat[Ours]{%
		\begin{minipage}[b]{0.20\linewidth}
			\includegraphics[width=1\linewidth]{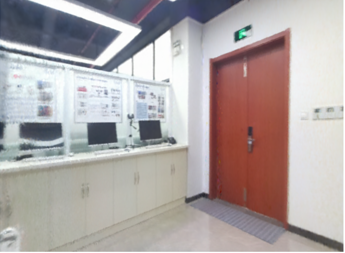}%
		\end{minipage}
	}
     \subfloat[Ground Truth]{%
		\begin{minipage}[b]{0.20\linewidth}
			\includegraphics[width=1\linewidth]{img/real_result/labor2/ours.png}%
		\end{minipage}
	}
	\caption{\textbf{Comparison of novel view rendering on the real-world dataset.} This is also supported by the quantitative results
    in Table~\ref{real data}.}
	\label{real render}
\end{figure*}

\textit{Real-world Dataset.}
A real-world dataset was created based on the rotating acquisition device proposed in \cite{10582466} used to capture common indoor scene information and compile it into a realistic scene dataset. The dataset includes three types of static scenes: offices, laboratories, and hotels, each captured from various positions with multiple sets of sensor information sequences. Each scene varies in size, encompassing challenging areas such as flat and low-texture regions. During data collection, the acquisition device was positioned at appropriate locations within the scenes, controlling the device to rotate at a constant speed exceeding 360°, approximately $0.25\pi$ radians per second.

\subsection{Baselines and Metrics.}
We primarily benchmark our SLAM method against existing state-of-the-art GS-based SLAM approaches such as SplaTAM~\cite{keetha2024splatam}, MonoGS~\cite{Matsuki:Murai:etal:CVPR2024}, Gaussian-SLAM~\cite{yugay2023gaussianslam}. SplaTAM achieves faster rendering and optimization by streamlining the dense mapping process, while MonoGS employs a co-visibility graph for optimization and Gaussian-SLAM organizes environments into sub-maps, optimizing them independently.

Following the evaluation metrics from GS-based SLAM SplaTAM~\cite{keetha2024splatam}, we employ standard photometric rendering quality metrics to assess the quality of novel view rendering, including Peak Signal-to-Noise Ratio (PSNR), Structural Similarity Index Measure (SSIM), and Learned Perceptual Image Patch Similarity (LPIPS). Furthermore, the quality of geometry generated by different approaches is evaluated by the L1 distance between reconstructed depth maps and their ground truth. Since the camera pose estimation performance is also a very important function for SLAM methods, we report the Root Mean Square Error (RMSE) of the Absolute Trajectory Error (ATE) of all frames.

\subsection{Novel View Rendering} \label{nvr}
As listed in Table~\ref{virtual w/o jitters} and~\ref{virtual}, approaches' rendering performances are tested in virtual sequences, where the difference between two types of sequence is that images and trajectories in Table~\ref{virtual w/o jitters} are added noise. What is more, the approaches tested are the original published version. Compared to MonoGS, Gaussian-SLAM, and SplaTAM, the proposed method has shown better rendering quality in all sequences in Tablee~\ref{virtual w/o jitters} and~\ref{virtual}. For example, in the  `office ($2^{\circ}$)' sequence, the PSNR result of the proposed method is $38.19$ dB, which is $3.3$ dB and $4.38$ dB higher than that of SplaTAM and MonoGS, respectively. Furthermore, from another perspective, in the same environment, like the `room' scene, but from different viewpoints, the proposed method has better performance in `room ($2^{\circ}$)'. Additionally, as the device rotation increases, the rendering performance of this method will also degrade from $38.19$ dB to $37.55$ dB since the overlaps between frames become larger incrementally. A similar phenomenon can be seen in Table~\ref{virtual}, the average rendering performance in PSNR, SSIM, and LPIPS are $38.678$ dB, $0.992$, and $0.026$, respectively, which show more accurate and robust performance compared to other approaches. Especially for MonoGS, there are 30\% performance drop from our $38.678$ dB to $28.422$ dB in the PSNR metric.  As shown in Figure~\ref{visual render}, the rendered images generated from different approaches in the `office' and `room' scenes are illustrated. For the results of MonoGS, it is easy to detect outlier areas. Limited by camera pose estimation, the 3D Gaussian maps of MonoGS have the problem of scene splitting, leading to the rendering performance in the first, second, and fourth rows.

Different from virtual sequences used in Table~\ref{virtual w/o jitters} and~\ref{virtual}, we evaluate these SLAM systems on real-world datasets as shown in Table~\ref{real data}. Compared to synthesis datasets, the novel view rendering tasks are more difficult in real-world sequences. In Figure~\ref{real render}, six scenes are illustrated. For MonoGS, the rendering results are very blurry, especially in `office\_0', `office\_1', and `office\_2' sequences. Compared to MonoGS, the Gaussian-SLAM method is more robust in these scenes, but it has worse results in indoor sequences, as shown in `labor\_0' and `labor\_1'. For the sequence `labor\_2', there is a black hole in the rendered image, where the area is a reflective area. However, in our method, the area is reconstructed correctly. Furthermore, compared to the results of SplaTAM and ground truth, it is easy to find that, there are alignment problems since the predicted camera poses and 3D Gaussians are not in the correct way.   

\subsection{Depth Estimation}
Based on 3D Gaussian models, we further render their depth maps to evaluate the quality of their 3D Gaussian models. By using an alpha-blending algorithm, the quality of the depth map becomes good when these Gaussian ellipsoids are aligned with the surface of the scenes.

As listed in Table~\ref{virtual w/o jitters} and \ref{virtual}, the proposed method has achieved robust depth rendering results in virtual sequences. To be specific, the average depth distance errors are $0.437$ cm and $0.586$ cm in virtual $-wo-noise$ and $-w-noise$, respectively. But, for other state-of-the-art approaches, the accuracy of depth maps is rendered in worse quality, where in virtual $-wo-noise$ sequences, the average results of MonoGS, Gaussian-SLAM, and SplaTAM are $2.75$ cm, $0.977$ cm, and $0.700$ cm. 

A similar phenomenon can be witnessed in Figure~\ref{diff depth}, where rendered depth maps generated virtual and real scenes are compared to ground truth depth maps. For MonoGS, the depth estimation performance in `office4n1' and `office\_0' sequences has a large distance. Compared to MnonGS, Gaussian-SLAM, and SplaTAM approaches show more robust results in most regions. But in some corner regions and low-textured regions, these two methods also suffer from high distance errors. For the proposed method, benefitting from the drift removal and optimization strategies,  more accurate depth estimation performance has been shown. 

\subsection{Pose Estimation} \label{VO}
Besides rendering performance in appearance and geometry, we also evaluate the pose estimation performance of these methods. As shown in Table~\ref{virtual w/o jitter ATE} and~\ref{virtual ATE}, the estimated trajectories are compared to the ground truth ones. Compared to these state of the art, the proposed method shows robust and accurate camera pose estimation results. From the perspective of average ATE errors, this method is 2 times more accurate than the Gaussian-SLAM method, and more accurate than SplaTAM and MonoGS. 

% 无抖动虚拟数据集ATE对比 
\begin{table*}[ht]
    \centering
    \renewcommand{\arraystretch}{1.2} %rows, default value is 1.0
    %\captionsetup[table*]{singlelinecheck=off}
	\setlength{\tabcolsep}{11pt}
    \resizebox{\textwidth}{!}{
    \begin{tabular}{c|ccccccc}
    \toprule
        Method  & \multicolumn{1}{l}{room ($2^{\circ}$)} & \multicolumn{1}{l}{room ($3^{\circ}$)} & \multicolumn{1}{l}{room ($4^{\circ}$)} & \multicolumn{1}{l}{office ($2^{\circ}$)} & \multicolumn{1}{l}{office ($3^{\circ}$)} & \multicolumn{1}{l}{office ($4^{\circ}$)} & \multicolumn{1}{l}{Avg.} \\
        \hline
        \multirow{1}{*}{MonoGS\cite{Matsuki:Murai:etal:CVPR2024}}  & 5.20 & 6.60   & 8.80  & 9.13   & 5.20  & 11.05  & 7.663 \\
        \hline
        \multirow{1}{*}{Gaussian-SLAM\cite{yugay2023gaussianslam}}  & 0.94 & 0.99   & 1.30& 1.75   & 1.17  & 0.92  & 1.178 \\
        \hline
        \multirow{1}{*}{SplaTAM\cite{keetha2024splatam}}  & 0.729 & 0.679   & 0.517  & 0.793   & 0.669  & 0.831  & 0.703 \\
        \hline
        \multirow{1}{*}{Ours}  &\textbf{0.212}    & \textbf{0.221}   & \textbf{0.194}     & \textbf{0.162}   & \textbf{0.140}  & \textbf{0.170}  & \textbf{0.183} \\
    \bottomrule
    \end{tabular}}
\caption{\textbf{Trajectory errors in ATE [cm]$\downarrow$ on virtual sequences without noise and jitters.} Results with best accuracy are highlighted by \textbf{bold} font.}
\label{virtual w/o jitter ATE}
\end{table*}

% 有抖动虚拟数据集ATE对比 
\begin{table*}[ht]
    \centering
    \renewcommand{\arraystretch}{1.2} %rows, default value is 1.0
    %\captionsetup[table*]{singlelinecheck=off}
	\setlength{\tabcolsep}{11pt}
    \resizebox{\textwidth}{!}{
    \begin{tabular}{c|ccccccc}
    \toprule
        Method  & \multicolumn{1}{l}{room2n1} & \multicolumn{1}{l}{room3n1} & \multicolumn{1}{l}{room4n1} & \multicolumn{1}{l}{office2n1} & \multicolumn{1}{l}{office3n1} & \multicolumn{1}{l}{office4n1} & \multicolumn{1}{l}{Avg.} \\
        \hline
        \multirow{1}{*}{MonoGS\cite{Matsuki:Murai:etal:CVPR2024}}  & 4.60 & 8.42   & 4.81  & 8.39   & 10.2  & 9.98  & 7.733 \\
        \hline
        \multirow{1}{*}{Gaussian-SLAM\cite{yugay2023gaussianslam}}  & 0.44 & 0.49   & 0.50& 0.70   & 0.73  & 0.79  & 0.608 \\
        \hline
        \multirow{1}{*}{SplaTAM\cite{keetha2024splatam}}  & 0.772 & 0.63   & 0.57  & 0.834   & 0.734  & 0.893  & 0.739 \\
        \hline
        \multirow{1}{*}{Ours}  & \textbf{0.226} & \textbf{0.253}   & \textbf{0.255}  & \textbf{0.164}   & \textbf{0.161}  & \textbf{0.173}  & \textbf{0.205} \\
    \bottomrule
    \end{tabular}}
\caption{\textbf{ Trajectory errors in ATE [cm]$\downarrow$ on virtual sequences with noise and jitters.} Results with best accuracy are highlighted by \textbf{bold} font.}
\label{virtual ATE}
\end{table*}

%\newpage 
\subsection{Ablation Studies} \label{other}
In Table~\ref{ablation room} and ~\ref{ablation office}, we ablate four aspects of our system:(1) the use of timesteps to differentiate between novel and previously encountered 3D Gaussian. (2) the use of a pose graph to optimize global pose during loop detection. (3) After the pose graph optimization is completed, the global Gaussian map subsequently undergoes an additional 3000 iterations of refinement using the loss function mentioned in~\ref{mapping}. (4) After the pose graph optimization is completed, the global Gaussian map is then refined using the method Gaussian update mentioned earlier in ~\ref{Loop Closure}. We do this using room3n1 and office3n1 of the virtual dataset.
After optimizing the global camera poses using pose graph optimization, the accuracy of camera tracking improved, but the overall rendering performance showed minimal change and even declined in the dataset room3n1. Long-term iterative optimization of the global map can enhance overall rendering performance. However, this optimization is not sufficiently robust and the improvements are limited.

% % 对比实验
\begin{table}[!t]
    \centering
    \Huge
    \renewcommand{\arraystretch}{1.5} %rows, default value is 1.0
    %\captionsetup[table*]{singlelinecheck=off}
	\setlength{\tabcolsep}{11pt}
    \resizebox{\columnwidth}{!}{
    \begin{tabular}{ccccccc}
    \toprule
        \begin{tabular}[c]{@{}c@{}}\textbf{Time-}\\[-2ex]\textbf{stamp}\end{tabular}  & \begin{tabular}[c]{@{}c@{}}\textbf{Pose graph}\\[-2ex]\textbf{optimization}\end{tabular} &  \begin{tabular}[c]{@{}c@{}}\textbf{Iterative}\\[-2ex]\textbf{optimization}\end{tabular} & \begin{tabular}[c]{@{}c@{}}\textbf{Gaussian}\\[-2ex]\textbf{updating}\end{tabular} &  \begin{tabular}[c]{@{}c@{}}\textbf{ATE}\\[-2ex][cm]$\downarrow$\end{tabular} & \begin{tabular}[c]{@{}c@{}}\textbf{PSNR}\\[-2ex][dB]$\uparrow$\end{tabular} & \begin{tabular}[c]{@{}c@{}}\textbf{Dep. L1}\\[-2ex][cm]$\downarrow$\end{tabular} \\
        \hline
        \XSolidBrush & \XSolidBrush & \XSolidBrush   & \XSolidBrush & 0.568   & 36.31  & 0.62  \\
        %\hline
        \CheckmarkBold  & \CheckmarkBold &  \XSolidBrush   &  \XSolidBrush& 0.285   & 36.01  & 0.83\\
        %\hline
        \CheckmarkBold & \CheckmarkBold & \CheckmarkBold   &  \XSolidBrush  & 0.260   & 37.68  & 0.54  \\
        %\hline
        \CheckmarkBold  &\CheckmarkBold    &  \XSolidBrush   & \CheckmarkBold     & \textbf{0.253}  & \textbf{39.57}  & \textbf{0.49}  \\
    \bottomrule
    \end{tabular}}
\caption{\textbf{Ablation Studies of using different modules of the proposed system in virtual sequence room3n1.}}
\label{ablation room}
\end{table}

\begin{table}[!t]
    \centering
    \Huge
    \renewcommand{\arraystretch}{1.5} %rows, default value is 1.0
    %\captionsetup[table*]{singlelinecheck=off}
	\setlength{\tabcolsep}{11pt}
    \resizebox{\columnwidth}{!}{
    \begin{tabular}{ccccccc}
    \toprule
        \begin{tabular}[c]{@{}c@{}}\textbf{Time-}\\[-2ex]\textbf{stamp}\end{tabular}  & \begin{tabular}[c]{@{}c@{}}\textbf{Pose graph}\\[-2ex]\textbf{optimization}\end{tabular} &  \begin{tabular}[c]{@{}c@{}}\textbf{Iterative}\\[-2ex]\textbf{optimization}\end{tabular} & \begin{tabular}[c]{@{}c@{}}\textbf{Gaussian}\\[-2ex]\textbf{updating}\end{tabular} &  \begin{tabular}[c]{@{}c@{}}\textbf{ATE}\\[-2ex][cm]$\downarrow$\end{tabular} & \begin{tabular}[c]{@{}c@{}}\textbf{PSNR}\\[-2ex][dB]$\uparrow$\end{tabular} & \begin{tabular}[c]{@{}c@{}}\textbf{Dep. L1}\\[-2ex][cm]$\downarrow$\end{tabular} \\
        \hline
        \XSolidBrush & \XSolidBrush & \XSolidBrush   & \XSolidBrush & 0.235   & 36.64  & 0.76  \\
        %\hline
        \CheckmarkBold  & \CheckmarkBold &  \XSolidBrush   &  \XSolidBrush& 0.162   & 36.69  & 0.79\\
        %\hline
        \CheckmarkBold & \CheckmarkBold & \CheckmarkBold   &  \XSolidBrush  & \textbf{0.160}   & 36.88  & 0.72  \\
        %\hline
        \CheckmarkBold  &\CheckmarkBold    &  \XSolidBrush   & \CheckmarkBold     & 0.161  & \textbf{38.03}  & \textbf{0.68}  \\
    \bottomrule
    \end{tabular}}
\caption{\textbf{ Ablation Studies of using different modules of the proposed system in virtual sequence office3n1.}}
\label{ablation office}
\end{table}

\section{Conclusion and Future Work}
In this paper, we propose a robust Gaussian Splatting SLAM system for rotating devices with multiple RGB-D cameras, which achieves accurate localization and photorealistic rendering performance based on the novel Gaussian-based loop closure module. 
% Contribution 1
In the loop detection step, we first label a timestamp to each Gaussian and categorize Gaussians as historical or novel groups based on its timestamp. By rendering different Gaussian types at the same viewpoint, we propose a loop detection strategy that considers co-visibility relationships and distinct rendering outcomes.
% Contribution 2
Based on the loop detection results, a loop closure optimization approach is proposed to remove camera pose drift and maintain the high quality of 3D Gaussian models. The approach uses a lightweight pose graph optimization algorithm to correct pose drift and updates Gaussians based on the optimized poses. Additionally, a bundle adjustment scheme refines camera poses using photometric and geometric constraints.

In future work, we will explore to update the proposed Gaussian Splatting SLAM system for dynamic scenes by using 4D Gaussian Splatting algorithms. Based on the detection of dynamic regions, the static areas will be fed to the proposed loop closure module, while leveraging smooth motion constraints for dynamic objects to achieve more accurate tracking and rendering in more general environments.

%\section*{Acknowledgement}
% This work is supported the Key R\&D Program of Zhejiang under Grant (2023C01044), the National Nature Science Foundation of China (62301198), the Provincial Universities of Zhejiang under Grants (GK239909299001-013), the National Natural Science Foundation of China (U22A2047, 61931008), the National Key R\&D Program of China under Grant (2023YFB4502800, 2023YFB4502803, 2020YFB1406604) and the Lishui Institute of Hangzhou Dianzi University Fundation (KY2023001, KY2023004).

% \newpage
\bibliographystyle{IEEEtran}
\bibliography{3dgsslam}

\vfill

\end{document}